\documentclass[letterpaper, 10 pt, conference]{ieeeconf}

\IEEEoverridecommandlockouts                              

\usepackage[utf8]{inputenc} 
\usepackage[T1]{fontenc}    
\usepackage{hyperref}       
\usepackage{url}            
\usepackage{booktabs}       
\usepackage{amsfonts}       
\usepackage{nicefrac}       
\usepackage{microtype}      

\usepackage{graphicx, subcaption} 
\usepackage{amsmath} 
\usepackage{amssymb}  
\usepackage{algorithm}
\usepackage[noend]{algpseudocode}

\usepackage{multirow}
\usepackage{lscape}
\usepackage{bbm}
\usepackage{bm}
\usepackage{amsmath}
\usepackage[normalem]{ulem}
\usepackage{textcomp}

\usepackage[colorinlistoftodos]{todonotes}

\renewcommand{\textuparrow}{$\uparrow$}
\renewcommand{\textdownarrow}{$\downarrow$}

\graphicspath{ {figures/} }

\DeclareMathOperator{\E}{\mathbb{E}}

\title{\LARGE \bf
Learning robust driving policies without online exploration
}

\author{
  Daniel Graves$^{1}$, Nhat M. Nguyen$^{1}$, Kimia Hassanzadeh$^{1}$, Jun Jin$^{1}$, Jun Luo$^{1}$
  \thanks{$^{1}$Noah's Ark Lab, Huawei Technologies Canada
    {\tt\small \{daniel.graves, minh.nhat.nguyen, kima.hassanzadeh, jun.jin1, jun.luo1\}@huawei.com}}
}

\begin{document}

\maketitle
\thispagestyle{empty}
\pagestyle{empty}

\begin{abstract}
We propose a multi-time-scale predictive representation learning method to efficiently learn robust driving policies in an offline manner that generalize well to novel road geometries, and damaged and distracting lane conditions which are not covered in the offline training data. We show that our proposed representation learning method can be applied easily in an offline (batch) reinforcement learning setting demonstrating the ability to generalize well and efficiently under novel conditions compared to standard batch RL methods. Our proposed method utilizes training data collected entirely offline in the real-world which removes the need of intensive online explorations that impede applying deep reinforcement learning on real-world robot training. Various experiments were conducted in both simulator and real-world scenarios for the purpose of evaluation and analysis of our proposed claims.  

\end{abstract}

\section{Introduction}


Learning to drive is a challenging problem that is a long-standing goal in robotics and autonomous driving.
In the early days of autonomous driving, a popular approach to staying within a lane was based on lane marking detection \cite{nikolaus2006lanekeeping}.
However, a significant challenge with this approach is the lack of robustness to missing, occluded or damaged lane markings \cite{qin2019robustlane} where most roads in the US are not marked with reliable lane markings on either side of the road \cite{ort2018}.
Modern approaches mitigate some of these issues by constructing high definition maps and developing accurate localization techniques \cite{feiwu2007, garimella2017, liu2020hdmapoverview, wang2017lidarlocalization}.
However, scaling both the map and localization approaches globally in a constantly changing world is still very challenging for autonomous driving and robotic navigation \cite{feiwu2007}.

Recently, there have been a growing number of successes in AI applied to robotics and autonomous driving \cite{chen2019imitation, bojarski2016, chen2017,sallab2017,chi2017,ort2018}.
These data-driven approaches can be divided into two categories: (1) behavior cloning, and (2) reinforcement learning (RL).
Behavior cloning suffers from generalization challenges since valuable negative experiences are rarely collected; in addition they cannot offer performance better than the behavior being cloned \cite{chen2019imitation,pan2020imitation,chi2017}.
RL on the other-hand is a promising direction for vision-based control \cite{shixiang2017}; however, RL is usually not practical because it requires extensive online exploration in the environment to find the best policy that maximizes the cumulative reward \cite{sallab2017,kiran2020deeprl,arnold2019realworldrl}.
Moreover, the success in game environments like Go \cite{silver2016} doesn't always transfer well to success in the real-world where an agent is expected to learn policies that generalize well \cite{akkaya2019solving, arnold2019realworldrl}.
A key challenge is that RL overfits to the training environment where learned policies tend to perform poorly on novel situations not seen during training  \cite{whiteson2011,zhao2019,henderson2017,farebrother2018}.
We aim to address the issues of learning both practical and general driving policies in the real-world by combining a novel representation learning approach with offline RL without any online exploration.

\begin{figure}[t]
\centering
\includegraphics[width=8cm]{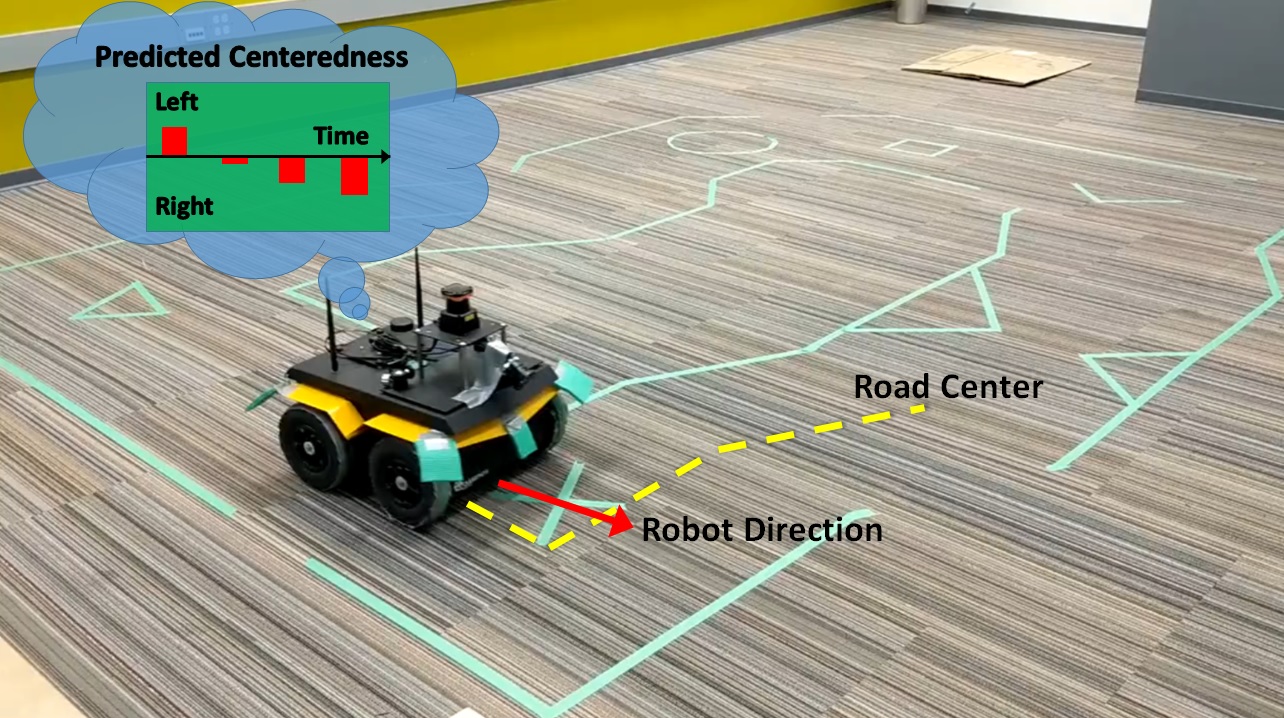}
\caption{Lane keeping of a Jackal Robot using vision-based counterfactual predictions of the future lane centeredness over multiple time scales to represent the state of the agent in RL}
\label{fig_jackal_overview}
\end{figure}

Offline RL, or learning RL policies from data without exploration \cite{fujimoto2018off}, could potentially address many of the practicality issues with applying standard online RL in the real-world.
Unfortunately, deep offline RL struggles to generalize to data not in the training set \cite{fujimoto2019benchmark,levine2020offline}.
Our approach applies novel representation learning based on counterfactual predictions \cite{sutton2011,modayil2012,levine2020offline}, shown in Figure  \ref{fig_jackal_overview}, to address the generalization issue.
We learn predictions of future lane centeredness and road angle from offline data safely collected in the environment using noisy localization sources during training, eliminating the need for expensive, on-vehicle, high accuracy localization sensors during deployment.
State-of-the-art offline RL \cite{fujimoto2018off} is then applied using these counterfactual predictions as a low dimensional representation of the state to learn a policy to drive the vehicle.
These counterfactual predictions are motivated in psychology \cite{clark2013,russek2017} where we find predictions aid in an agent's understanding of the world, particularly in driving \cite{salvucci2004}.
Similar works in classical control have shown how anticipation of the future is important for driving at the limits of stability through feed-forward \cite{kapania2015feedforwardlimits} and model predictive control \cite{beal2013mpclimits}.
Our work is motivated by the predictive state hypothesis \cite{littman2002,rafols2005} that claims counterfactual predictions help an agent generalize and adapt quickly to new problems \cite{schaul2013}.

The significance of our approach is that it demonstrates practical value in autonomous driving and real-world RL without requiring extensive maps, robust localization techniques or robust lane marking and curb detection.
We demonstrate that our approach generalizes to never-before seen roads including those with damaged and distracting lane markings.
The novel contributions of this work are summarized as follows: (1) an algorithm for learning counterfactual predictions from real-world driving data with behavior distribution estimation, (2) an investigation into the importance of predictive representations for learning good driving policies that generalize well to new roads and damaged lane markings, and (3) the first demonstration of deep RL applied to autonomous driving with real-world data without any online exploration.

\section{Related Works}
\textbf{Deep learning approaches to driving}: There have been many attempts to apply deep learning to driving including deep RL and imitation learning \cite{sallab2017}; however generalization is a key challenge.
ChaufferNet \cite{bansal2019chauffeurnet} used a combination of imitation learning and predictive models to synthesize the worst case scenarios but more work is needed to improve the policy to achieve performance competitive with modern motion planners.
Another approach trained the agent entirely in the simulator where transfer to the real-world could be challenging to achieve \cite{sallab2017}.
DeepDriving \cite{chen2015} learned affordance predictions of road angle from an image for multi-lane driving in simulation using offline data collected by human drivers.
However, in contrast with our proposed method, DeepDriving used heuristics and rules to control the vehicle instead of learning a policy with RL.
Moreover, DeepDriving learned predictions of the current lane centeredness and current road angle rather than long-term counterfactual predictions of the future.

\textbf{Offline RL in real-world robot training}: There are many prior arts in offline (batch) RL \cite{thomas2016,fujimoto2019benchmark}.
However, most prior arts in offline RL have challenges learning good policies in the deep setting \cite{fujimoto2019benchmark}.
The current state of the art in offline RL is batch constrained Q-learning (BCQ) \cite{fujimoto2018off,fujimoto2019benchmark} where success is demonstrated in simulation environments such as Atari but the results still perform badly in comparison to online learning.
The greatest challenge with offline RL is the difficulty in covering the state-action space of the environment resulting in holes in the training data where extrapolation is necessary.
\cite{cunha2015batchrl_soccer} applied a novel offline RL approach to playing soccer with a real-world robot by exploiting the episodic nature of the problem.
Our work overcomes these challenges and is, to the best of our knowledge, the first successful real-world robotic application of batch RL with deep learning.

\textbf{Counterfactual prediction learning}: 
Learning counterfactual predictions as representation of the state of the agent has been proposed before in the real-world \cite{gunther2016,edwards2016}.
Other approaches demonstrate counterfactual predictions but don't provide a way to use them \cite{sutton2011,awhite2015,modayil2012}.
While experiments with counterfactual predictions show a lot of promise for improving learning and generalization, most experiments are in simple tabular domains \cite{littman2002,rafols2005,schaul2013}.
Auxiliary tasks and similar prediction problems have been applied to deep RL task in simulation but assume the policy is the same as the policy being learned and thus are not counterfactual predictions \cite{jaderberg2018,barreto2017,russek2017,vanseijen2017}.

\section{Predictive Control for Autonomous Driving}
Let us consider the usual setting of an MDP described by a set of states $S$, a set of actions $A$, and transition dynamics with probability $P(s'|s,a)$ of transitioning to next state $s'$ after taking action $a$ from state $s$, and a reward $r$.
The objective of an MDP is to learn a policy $\pi$ that maximizes the future discounted sum of rewards in a given state.
Obtaining the state of the agent in an MDP environment is not trivial especially with deep RL where the policy is changing because the target is moving \cite{minh2013}.
Our approach is to learn an intermediate representation mapping sensor readings $s$ to a limited number of counterfactual predictions $\phi$ as a representation of the state for deep RL.
This has the advantage of pushing the heavy burden of deep feature representation learning in RL to the easier problem of prediction learning \cite{schlegel2019,ghiassian2018,graves2019,jaderberg2018}.

\begin{figure}[t]
\centering
\includegraphics[width=8cm]{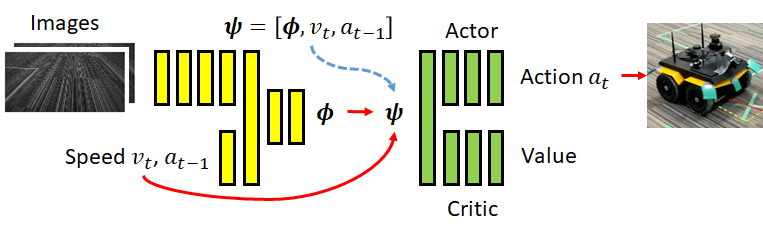}
\caption{Overall architecture of the RL system involved learns a predictive representation $\bm{\psi}$ to represent the state of the agent. Camera is the only environment sensor at test time.}
\label{fig_overall_architecture}
\end{figure}

The overall architecture of the system is depicted in Figure \ref{fig_overall_architecture}.
The proposal is to represent the state of the agent as a vector $\bm{\psi}$ which is the concatenation of a limited number of the predictions $\bm{\phi}$, the current speed of the vehicle $v_t$ and the previous action taken $a_{t-1}$.
The predictions $\bm{\phi}$ are counterfactual predictions, also called general value functions \cite{sutton2011}.
The previous action taken is needed due to the nature of the predictions which are relative to the last action.

Learning a policy $\pi(\bm{\psi})$ could provide substantial benefits over learning $\pi$ from image observations: (1) improving learning performance and speed, (2) enabling batch RL from offline data, and (3) improving generalization of the driving policy.
Our approach is to learn a value function $Q(s,a)$ and a deterministic policy $\pi(\bm{\psi})$ that maximizes that value function using batch constrained Q-learning (BCQ) \cite{fujimoto2018off}.
While the networks can be modelled as one computational graph, the gradients from the policy and value function network are not back-propagated through the prediction network to decouple the representation learning when learning from the offline data.
Thus, training happens in two phases: (1) learning the prediction network, (2) learning the policy and value function.

\begin{figure}[t]
	\centering
	\begin{subfigure}[b]{4cm}
		\includegraphics[width=4cm]{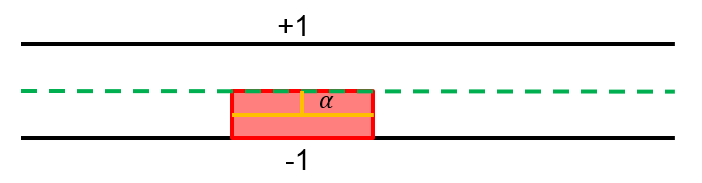}
		\caption{Lane centeredness $\alpha$}
		\label{fig_trackpos}
	\end{subfigure}
	\begin{subfigure}[b]{4cm}
		\includegraphics[width=4cm]{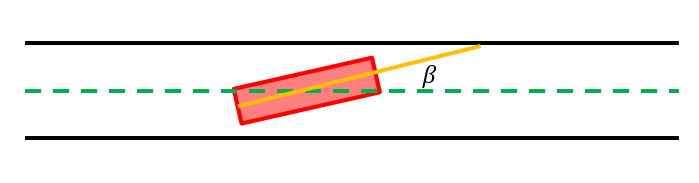}
		\caption{Road angle $\beta$}
		\label{fig_roadangle}
	\end{subfigure}
\caption{An illustration of (a) lane centeredness position $\alpha$, and (b) the road angle $\beta$ which is the angle between the direction of the vehicle and the direction of the road.}
\label{fig_road_cumulants}
\end{figure}
\begin{figure}[t]
	\centering
	\includegraphics[width=7cm]{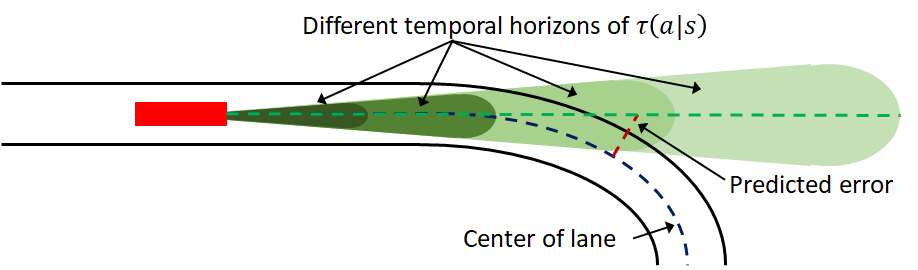}
    \caption{An illustration of the multiple temporal horizons of the predictions $\phi$.}
\label{fig_road_predictions}
\end{figure}

During the first phase of training, a low-accuracy localization algorithm, based on 2D lidar scan matching, produces the lane centeredness $\alpha$ and relative road angle $\beta$ of the vehicle, depicted in Figure \ref{fig_road_cumulants}, that are used to train the prediction network.
The prediction network is a single network that predicts the lane centeredness and relative road angle over multiple temporal horizons depicted in Figure \ref{fig_road_predictions}:  these are predictions of the future lane centeredness and relative road angle rather than the current estimates returned by the localization algorithm.
They are chosen because they represent both present and future lane centeredness information needed to steer \cite{salvucci2004}.
These predictions are discounted sums of future lane centeredness and relative road angle respectively that are learned with GVFs \cite{sutton2011}:

\begin{equation}
    \bm{\phi}(s)=\E_{\tau}[\sum_{i=0}^{\infty}{\gamma^i \bm{c}_{t+i+1}} | s_t=s]
    \label{eq_predictive_representation}
\end{equation}

\noindent
where $\bm{c}_{t+i+1}$ is the cumulant vector consisting of the current lane centeredness $\alpha$ and current relative road angle $\beta$.
It is important to understand that $\bm{\phi}(s)$ predicts the sum of all future lane centeredness and road angle values collected under some policy $\tau$.
The policy $\tau$ is counterfactual in the sense that it is different from the behavior policy $\mu$ used to collect the data and the learned policy $\pi$.
Formally, the policy $\tau(a_t|s_t, a_{t_1})=\mathcal{N}(a_{t-1}, \Sigma)$ where $\Sigma=0.0025I$ is a diagonal covariance matrix.
The meaning of this policy is to ``keep doing what you are doing'', similar to the one used in \cite{beal2013mpclimits} for making counterfactual predictions.
Therefore, $\bm{\phi}(s)$ predicts the discounted sum of future lane centeredness and road angle if the vehicle takes similar actions to its last action.
Moreover, $\bm{\phi}(s)$ can be interpreted as predictions of the deviation from the desired lane centeredness and road angle.
Counterfactual predictions can be thought of as anticipated future "errors" that allow controllers to take corrective actions before the errors occur.
The discount factor $\gamma$ controls the temporal horizon of the prediction.
It is critical to learn $\bm{\phi}(s)$ for different values of $\gamma$ in order to control both steering and speed.
The details for learning $\bm{\phi}(s)$ are provided in the next section.

During the second stage of training, the localization algorithm is no longer needed; it was used to provide the labels for training the predictive representation in the first stage.
Instead, the counterfactual predictions $\bm{\phi}$ are concatenated with the vehicle speed $v_t$ and last action $a_{t-1}$ to form a predictive representation $\bm{\psi}$.
The RL agent receives $\bm{\psi}$ as the state of the agent in the environment which is used to predict the value and produce the next action $a_t$ as depicted in Figure \ref{fig_overall_architecture}.
In our offline learning approach, we used the state of the art batch RL BCQ \cite{fujimoto2018off}\cite{fujimoto2019benchmark} to train the policy.

Note that the same architecture can also be applied online where the counterfactual prediction, policy and value networks are all learned online simultaneously with deep deterministic policy gradient (DDPG) \cite{silver2014} but the details are left in the appendix.


\section{Predictive Learning}
The counterfactual predictions given in Equation \eqref{eq_predictive_representation} are general value functions (GVFs) \cite{sutton2011} that are learned with a novel combination of different approaches including (1) off-policy, or counterfactual, prediction learning with importance resampling \cite{schlegel2019}, and (2) behavior estimation with the density ratio trick \cite{sugiyama2010densityratio}.
\subsection{Counterfactual Predictions}
To ask a counterfactual predictive question, we use the GVF framework, where one must define a cumulant $c_t=c(s_t,a_t,s_{t+1})$, a.k.a. pseudo-reward, a policy distribution $\tau(a|s)$ and continuation function $\gamma_t=\gamma(s_t,a_t,s_{t+1})$.
The answer to the predictive question is the expectation of the return $\phi_t$, when following policy $\tau$, defined by

\begin{equation}
    \phi^\tau(s) = \E_{\tau}[\sum_{k=0}^{\infty} {(\prod_{j=0}^{k-1} {\gamma_{t+j+1}}) c_{t+k+1}} | s_t=s,a_t=a]
\label{eq_value}
\end{equation}

\noindent
where the cumulant is $c_t$ and $0 \leq \gamma_t \leq 1$ \cite{sutton2011}.
This is the more general form for learning a prediction than the one given in Equation \eqref{eq_predictive_representation} where the only difference is that $\gamma$ is replaced by a continuation function which allows for predictions that predict the sum of cumulants until an episodic event occurs such as going out of lane.
The agent usually collects experience under a different behavior policy $\mu(a|s)$.
When $\tau$ is different from both the behavior policy $\mu$ and the policy being learned $\pi$, the predictive question is a counterfactual prediction\footnote{Some literature call this an off-policy prediction}.
Cumulants are often scaled by a factor of $1-\gamma$ when $\gamma$ is a constant in non-episodic predictions.
The counterfactual prediction $\phi^\tau(s)$ is a general value function (GVF) is approximated by a deep neural network parameterized by $\theta$ to learn \eqref{eq_value}.
The parameters $\theta$ are optimized with gradient descent minimizing the following loss function

\begin{equation}
L(\theta) = \E_{\mu}[\rho \delta^2]
\label{eq_td_loss}
\end{equation}

\noindent
where $\delta=\phi^\tau(s;\theta) - y$ is the TD error and $\rho=\frac{\tau(a|s)}{\mu(a|s)}$ is the importance sampling ratio to correct for the difference between the policy distribution $\tau$ and behavior distribution $\mu$.
Note that only the behavior policy distribution is corrected; but the expectation is still over the state visitation distribution under the policy $\mu$.
In practice, this is usually not an issue \cite{schlegel2019}.
The target $y$ is produced by bootstrapping a prediction of the value of the next state \cite{sutton1988} under policy $\tau$ 

\begin{equation}
y=\E_{s_{t+1} \sim P}[c_{t+1} + \gamma \phi^\tau(s_{t+1};\hat{\theta})|s_t=s,a_t=a]
\label{eq_td_bootstrap}
\end{equation}

\noindent
where $y$ is a bootstrapped prediction using the most recent parameters $\hat{\theta}$ that are assumed constant in the gradient computation.
Learning a counterfactual prediction with a fixed policy $\tau$ tends to be very stable when minimizing $L(\theta)$ using gradient descent approaches and therefore doesn't require target networks originally used in \cite{minh2013} to stabilize DQN.

The gradient of the loss function \eqref{eq_td_loss} is given by
\begin{equation}
\nabla_{\theta} L(\theta) = \E_{\mu}[\rho \delta \nabla_{\theta} \phi^\tau(s;\theta)]
\label{eq_td_is_gradient}
\end{equation}

\noindent
However, updates with importance sampling ratios are known to have high variance which may negatively impact learning; instead we use the importance resampling technique to reduce the variance of the updates \cite{schlegel2019}.
With importance resampling, a replay buffer $D$ of size $N$ is required and the gradient is estimated from a mini-batch and multiplied with the average importance sampling ratio of the samples in the buffer $\bar{\rho}=\frac{\sum_{i=1}^{N}{\rho_i}}{N}$.

The gradient with importance resampling is given by

\begin{equation}
\nabla_{\theta} L(\theta) = \E_{s,a \sim D_{\rho}}[\bar{\rho} \delta \nabla_{\theta} \hat{v}^\tau(s;\theta)]
\label{eq_td_ir_gradient}
\end{equation}

\noindent
where $D_{\rho}$ is a distribution of the transitions in the replay buffer proportional to the importance sampling ratio.
The probability for transition $i=1...N$ is given by $D_i=\frac{\rho_i}{\sum_{j=1}^{N}{\rho_j}}$ where the importance sampling ratio is $\rho_i=\frac{\tau(a_i|s_i)}{\mu(a_i|s_i)}$.
An efficient data structure for the replay buffer is the SumTree used in prioritized experience replay \cite{schaul2016}.

\subsection{Behavior Estimation}
When learning predictions from real-world driving data, one needs to know the behavior policy distribution $\mu(a|s)$; however, in practice this is rarely known.
Instead we estimate it using the density ratio trick \cite{sugiyama2010densityratio} where the ratio of two probability densities can be expressed as a ratio of discriminator class probabilities that distinguish samples from the two distributions.
Let us define an intermediate probability density function $\eta(a|s)$ such as the uniform distribution; this will be compared to the behavior distribution $\mu(a|s)$ which we desire to estimate.
The class labels $y=+1$ and $y=-1$ are labels given to samples from $\mu(a|s)$ and $\eta(a|s)$.
A discriminator $g(a,s)$ is learned that distinguishes state action pairs from the two distributions using the cross-entropy loss.
The ratio of the densities can be computed using only the discriminator $g(a,s)$.

\begin{equation}
    \begin{split}
        \frac{\mu(a|s)}{\eta(a|s)} & = \frac{p(a|s,y=+1)}{p(a|s,y=-1)} = \frac{p(y=+1|a,s)/p(y=+1)}{p(y=-1|a,s)/p(y=-1)} \\
            & = \frac{p(y=+1|a,s)}{p(y=-1|a,s)} = \frac{g(a,s)}{1 - g(a,s)}
    \end{split}
\label{eq_condition_density_ratio}
\end{equation}

Here we assume that $p(y=+1)=p(y=-1)$.
From this result, we can estimate $\mu(a|s)$ with $\hat{\mu}(a|s)$ as follows

\begin{equation}
\hat{\mu}(a|s) = \frac{g(a,s)}{1 - g(a,s)} \eta(a|s)
\label{eq_behavior_mu}
\end{equation}

\noindent
where $\eta(a|s)$ is a known distribution over action conditioned on state.
Choosing $\eta(a|s)$ to be the uniform distribution ensures that the discriminator is well trained against all possible actions in a given state; thus good performance is achieved with sufficient coverage of the state space rather than the state-action space.
Alternatively, one can estimate the importance sampling ratio without defining an additional distribution $\eta$ by replacing the distribution $\eta$ with $\tau$; however, defining $\eta$ to be a uniform distribution ensures the discriminator is learned effectively across the entire action space.
The combined algorithms for training counterfactual predictions with an unknown behavior distribution are given in the Appendix for both the online and offline RL settings.

\section{Experiments}
Our approach to learning counterfactual predictions for representing the state used in RL to learn a driving policy is applied to two different domains.
The first set of experiments is conducted on a Jackal robot in the real-world where we demonstrate the practicality of our approach and its robustness to damaged and distracting lane markings.
The second set of experiments is conducted in the TORCS simulator where we conduct an ablation study to understand the effect different counterfactual predictive representations have on performance and comfort.
Refer to the Appendix\footnote{Appendix is at https://bit.ly/3mIDScp} for more details in the experimental setup and training.

\subsection{Jackal Robot}
The proposed solution for learning to drive the Jackal robot in the real-world is called GVF-BCQ since it combines our novel method of learning GVF predictions with BCQ \cite{fujimoto2018off}.
Two baselines are compared with our method: (1) a classical controller using model predictive control (MPC), and (2) batch-constrained Q-learning that trains end-to-end (E2E-BCQ).
The MPC uses a map and 2D laser scanner for localization from pre-existing ROS packages.
The E2E-BCQ is the current state-of-the-art in offline deep RL \cite{fujimoto2019benchmark}.
Comparing to online RL was impractical for safety concerns and the need to recharge the robot's battery every 4 hours.

The training data consisted of 6 training roads in both counter clock-wise (CCW) and clock-wise (CW) directions and 3 test roads where each of the 3 test roads had damaged variants.
All training data was flipped to simulate travelling in the reverse direction and balance the data set in terms of direction.
The training data was collected using a diverse set of drivers including human drivers by remote control and a pure pursuit controller with safe exploration; thus, the training data was not suitable for imitation learning.
The test roads were different from the training data: (1) a rectangle-shaped road with rounded outer corners, (2) an oval-shaped road, and (3) a complex road loop with many turns significantly different from anything observed by the agent during training.
In addition, the test roads included variants with damaged lane markings.
The reward is given by $r_t=v_t (\cos{\beta_t} + |\alpha_t|)$ where $v_t$ is the speed of the vehicle in km/h, $\beta_t$ is the angle between the road direction and the vehicle direction, and $\alpha_t$ is the lane centeredness.

\begin{table}[t]
\centering
\caption{Comparison of GVF-BCQ (our method) and E2E-BCQ (baseline) on Rectangle test road with 0.4 m/s target speed in both the CW and CCW directions.  GVF-BCQ exceeds performance of E2E-BCQ in all respects with higher overall speed, and far fewer out of lane events.  E2E-BCQ was deemed unsafe for further experiments.}
\label{table_summary_jackal_result_gvf_vs_e2e}
\begin{tabular}{|l|l|l|l|l|l|l|l|} 
\hline
\multicolumn{1}{|c|}{\begin{tabular}[c]{@{}c@{}}Method\end{tabular}} & \multicolumn{1}{c|}{\begin{tabular}[c]{@{}c@{}}Dir.\end{tabular}} & \multicolumn{1}{c|}{\begin{tabular}[c]{@{}c@{}}r/s\\\textuparrow \end{tabular}} & \multicolumn{1}{c|}{\begin{tabular}[c]{@{}c@{}}Speed\\\textuparrow \end{tabular}} & \multicolumn{1}{c|}{\begin{tabular}[c]{@{}c@{}}Off-\\center\\\textdownarrow~\end{tabular}} & \multicolumn{1}{c|}{\begin{tabular}[c]{@{}c@{}}Off-\\angle\\\textdownarrow \end{tabular}} & \multicolumn{1}{c|}{\begin{tabular}[c]{@{}c@{}}Out of\\Lane \textdownarrow \end{tabular}} \\ 
\hline
GVF-BCQ & CCW & \textbf{2.68 } & \textbf{0.32 } & \textbf{0.14 } & \textbf{0.13 } & \textbf{0.0\% } \\
E2E-BCQ & CCW & 1.26 & 0.18 & 0.26 & 0.24 & 3.8\% \\ 
\hline
GVF-BCQ & CW & \textbf{2.29} & \textbf{0.31} & \textbf{0.22} & \textbf{0.16} & \textbf{0.0\%} \\
E2E-BCQ\footnotemark & CW & -0.13 & 0.17 & 0.99 & 0.30 & 54.2\% \\
\hline
\end{tabular}
\end{table}

\footnotetext{E2E-BCQ failed to recover after undershooting the first turn in the clock-wise (CW) direction; it was not safe for testing on the other roads.}

\begin{table}[t]
\centering
\caption{Effect of damaged lanes on GVF-BCQ performance in CCW direction with 0.4 m/s target speed where R, O, and C are the Rectangle, Oval and Complex road shapes respectively.  GVF-BCQ demonstrates robustness to damaged and distracting lanes.}
\label{table_summary_jackal_result_N_vs_D}
\begin{tabular}{|l|l|l|l|l|l|l|l|} 
\hline
 & \multicolumn{1}{|c|}{\begin{tabular}[c]{@{}c@{}}Damage \end{tabular}} & \multicolumn{1}{c|}{\begin{tabular}[c]{@{}c@{}}r/s \textuparrow \end{tabular}} & \multicolumn{1}{c|}{\begin{tabular}[c]{@{}c@{}}Off-\\center\\\textdownarrow \end{tabular}} & \multicolumn{1}{c|}{\begin{tabular}[c]{@{}c@{}}Off-\\angle\\\textdownarrow \end{tabular}} & \multicolumn{1}{c|}{\begin{tabular}[c]{@{}c@{}}Out of\\Lane \textdownarrow \end{tabular}} & \multicolumn{1}{c|}{\begin{tabular}[c]{@{}c@{}}Speed\\Jerk\\\textdownarrow \end{tabular}} & \multicolumn{1}{c|}{\begin{tabular}[c]{@{}c@{}}Steer\\Jerk\\\textdownarrow \end{tabular}}  \\ 
\hline
\multirow{2}{*}{\begin{tabular}[c]{@{}l@{}}R\end{tabular}}
 & No & 2.68 & 0.13 & 0.13 & 0.0\% & 0.036 & 0.23 \\
 & Yes & \textbf{2.74 } & 0.14 & 0.14 & 0.0\% & -0.038 & 0.23 \\ 
\hline
\multirow{2}{*}{O}
 & No & \textbf{2.40 } & \textbf{0.28 } & 0.21 & \textbf{1.5\% } & 0.035 & 0.22 \\
 & Yes & 2.07 & 0.33 & 0.21 & 7.19\% & 0.033 & 0.21 \\ 
\hline
\multirow{2}{*}{\begin{tabular}[c]{@{}l@{}} C\end{tabular}}
 & No & \textbf{2.35 } & \textbf{0.22 } & \textbf{0.18 } & \textbf{0.0\% } & \textbf{0.034 } & \textbf{0.23 } \\
 & Yes & 2.11 & 0.31 & 0.24 & 9.42\% & 0.044 & 0.29 \\
\hline
\end{tabular}
\end{table}

\begin{table}[t]
\centering
\caption{Comparison of GVF-BCQ (our method) and MPC (baseline) in CCW direction with 0.4 m/s target speed where R, O, and C are the Rectangle, Oval and Complex road shapes respectively.}
\label{table_summary_jackal_result_gvf_vs_mpc}
\begin{tabular}{|l|l|l|l|l|l|l|l|} 
\hline
 & \multicolumn{1}{|c|}{\begin{tabular}[c]{@{}c@{}}Method \end{tabular}} & \multicolumn{1}{|c|}{\begin{tabular}[c]{@{}c@{}}r/s\\\textuparrow \end{tabular}} & \multicolumn{1}{c|}{\begin{tabular}[c]{@{}c@{}}Off-\\center\\\textdownarrow\end{tabular}} & \multicolumn{1}{c|}{\begin{tabular}[c]{@{}c@{}}Off-\\angle\\\textdownarrow \end{tabular}} & \multicolumn{1}{c|}{\begin{tabular}[c]{@{}c@{}}Out of\\Lane \textdownarrow \end{tabular}} & \multicolumn{1}{c|}{\begin{tabular}[c]{@{}c@{}}Speed\\Jerk\\\textdownarrow \end{tabular}} & \multicolumn{1}{c|}{\begin{tabular}[c]{@{}c@{}}Steer\\Jerk\\\textdownarrow \end{tabular}} \\
\hline
\multirow{2}{*}{\begin{tabular}[c]{@{}l@{}}R \end{tabular}}
 & GVF-BCQ & \textbf{2.68} & \textbf{0.13} & \textbf{0.13} & \textbf{0.0\%} & \textbf{0.036} & \textbf{0.23} \\
 & MPC & 0.97 & 0.53 & 0.19 & 20.4\% & 0.083 & 1.25 \\ 
\hline
\multirow{2}{*}{\begin{tabular}[c]{@{}l@{}}O \end{tabular}}
 & GVF-BCQ & \textbf{2.40} & \textbf{0.28} & 0.21 & \textbf{1.45\%} & \textbf{0.035} & \textbf{0.22} \\
 & MPC & 0.89/s & 0.53 & \textbf{0.20} & 22.7\% & 0.103 & 1.41 \\ 
\hline
\multirow{2}{*}{\begin{tabular}[c]{@{}l@{}}C \end{tabular}}
 & GVF-BCQ & \textbf{2.35} & \textbf{0.22} & \textbf{0.18} & \textbf{0.0\%} & \textbf{0.034} & \textbf{0.23} \\
 & MPC & 0.72  & 0.64 & 0.21 & 38.9\% & -0.063 & -1.21 \\ 
\hline
\end{tabular}
\end{table}

A comparison of the learned approaches is given in Table \ref{table_summary_jackal_result_gvf_vs_e2e} where GVF-BCQ approach exceeds the performance of E2E-BCQ in all respects demonstrating better performance at nearly double the speed.
Both GVF-BCQ and E2E-BCQ were trained with the same data sets and given 10M updates each for a fair comparison.
For GVF-BCQ, the first 5M updates were used for learning the counterfactual predictions and the second 5M updates were used for learning the policy from the predictive representation with BCQ.
They both received the same observations consisting of two stacked images, current vehicle speed, and last action and produced desired steering angle and speed.

GVF-BCQ was tested on roads with damaged and distracting lane markings as shown in Table \ref{table_summary_jackal_result_gvf_vs_mpc}.  The damaged and distracting lane markings for the complex test road loop are shown in Figure \ref{fig_jackal_overview}.
These results demonstrate robustness because the training data did not include roads with damaged or distracting lane markings.

GVF-BCQ was also compared to MPC in Table \ref{table_summary_jackal_result_gvf_vs_mpc} where GVF-BCQ was found to produce superior performance in nearly all metrics at a high target speed of 0.4 m/s.
The MPC performed poorly since it was difficult to tune for 0.4 m/s; performance was more similar at 0.25 m/s speeds where results are in the Appendix.
A clear advantage of GVF-BCQ is the stability and smoothness of control achieved at the higher speeds.

\subsection{Ablation Study in TORCS}
In order to understand the role of counterfactual predictions in representing the state of the agent, we conduct an ablation study in the TORCS simulator.
We compare representations consisting of future predictions at multiple time scale, future predictions at a single time scale and predictions with supervised regression of the current (non-future) lane centeredness and relative road angle.
These experiments were conducted with online RL using deep deterministic policy gradient (DDPG) \cite{silver2014} in order to more easily understand the impact of the different state representations on the learning process.

Our method is called GVF-DDPG and uses multiple time scales specified by the values $\gamma=[0.0, 0.5, 0.9, 0.95, 0.97]$.
Two variants of our method called GVF-0.95-DDPG and GVF-0.0-DDPG were defined to investigate the impact of different temporal horizons on performance, where $\gamma=0.95$ and $\gamma=0.0$ respectively.
It is worth pointing out that when $\gamma=0$, the prediction is myopic meaning that it reduces to a standard supervised regression problem equivalent to the predictions learned in \cite{chen2015}.
These methods receive a history of two images, velocity and last action and produce desired steering angle and vehicle speed action commands.

Some additional baselines include a kinematic-based steering approach based on \cite{paden2016} and two variants of DDPG with slightly different state representations.
The kinematic-based steering approach is treated as a "ground truth" controller since it has access to perfect localization information to steer the vehicle; unlike our approach, the speed is controlled independently.
The variants of DDPG are called (1) DDPG-Image and (2) DDPG-LowDim.
DDPG-Image is given a history of two images, velocity and last action while DDPG-LowDim is given a history of two images, velocity, last action, current lane centeredness $\alpha$ and relative road angle $\beta$ in the observation.
Both DDPG-Image and DDPG-LowDim output steering angle and vehicle speed action commands.
The performance of DDPG-LowDim serves as an ideal learned controller since it learns from both images and the perfect localization information.

\begin{figure}[t]
	\centering
	\includegraphics[width=7.5cm]{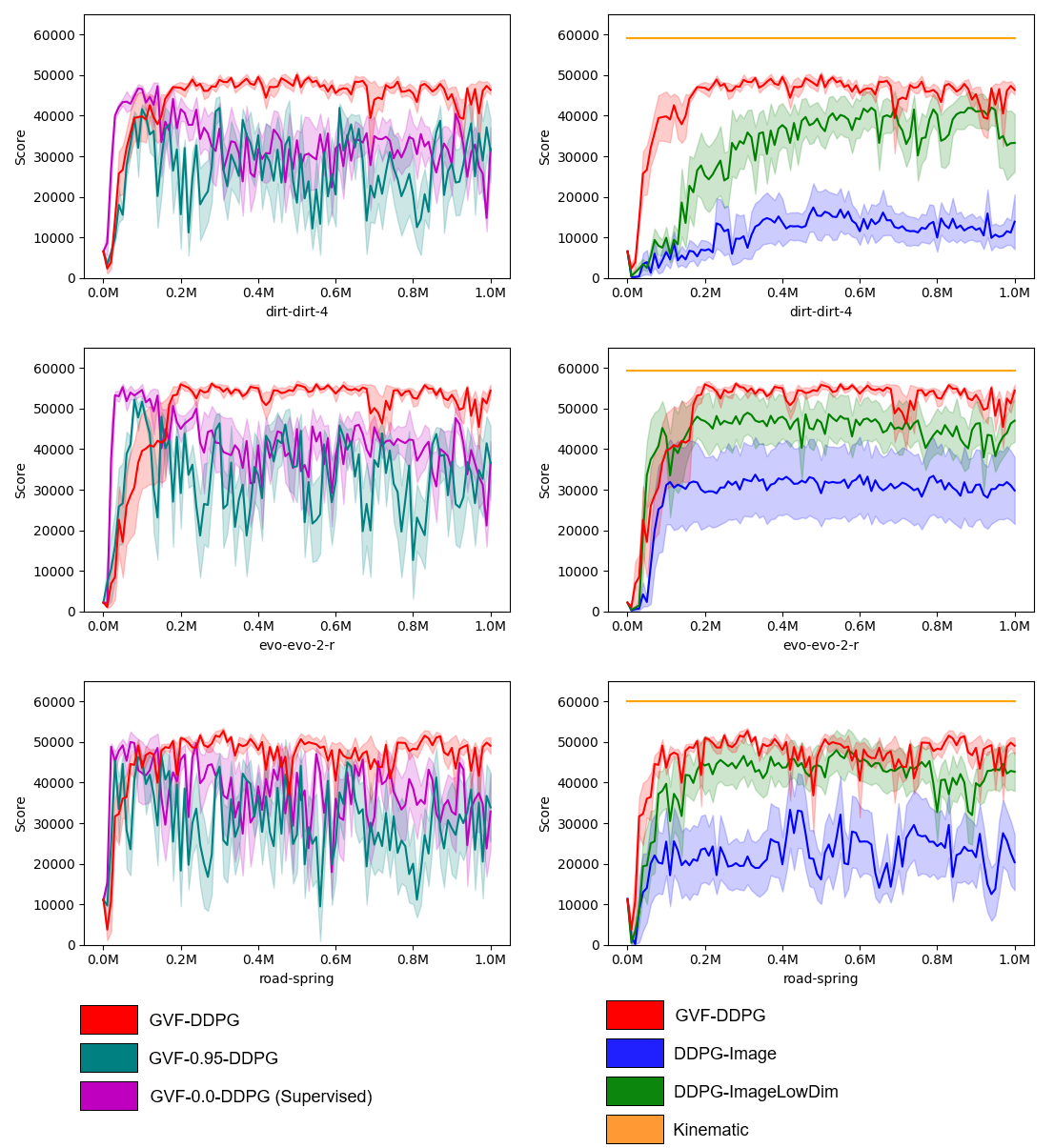}
	\caption{Ablation study of GVF-DDPG (our method) of test scores (accumulated reward) over different time scale selections (left) and raw image-based state representations (right). Test scores were evaluated every 1000 steps during training for dirt-dirt-4, evo-evo-2 and road-spring which were not part of the training set. Results show our proposed predictive representation with multiple time scales achieves the best performance.}
	\label{fig_summary_scores}
\end{figure}

\begin{figure}[t]
	\centering
	\includegraphics[width=7.5cm]{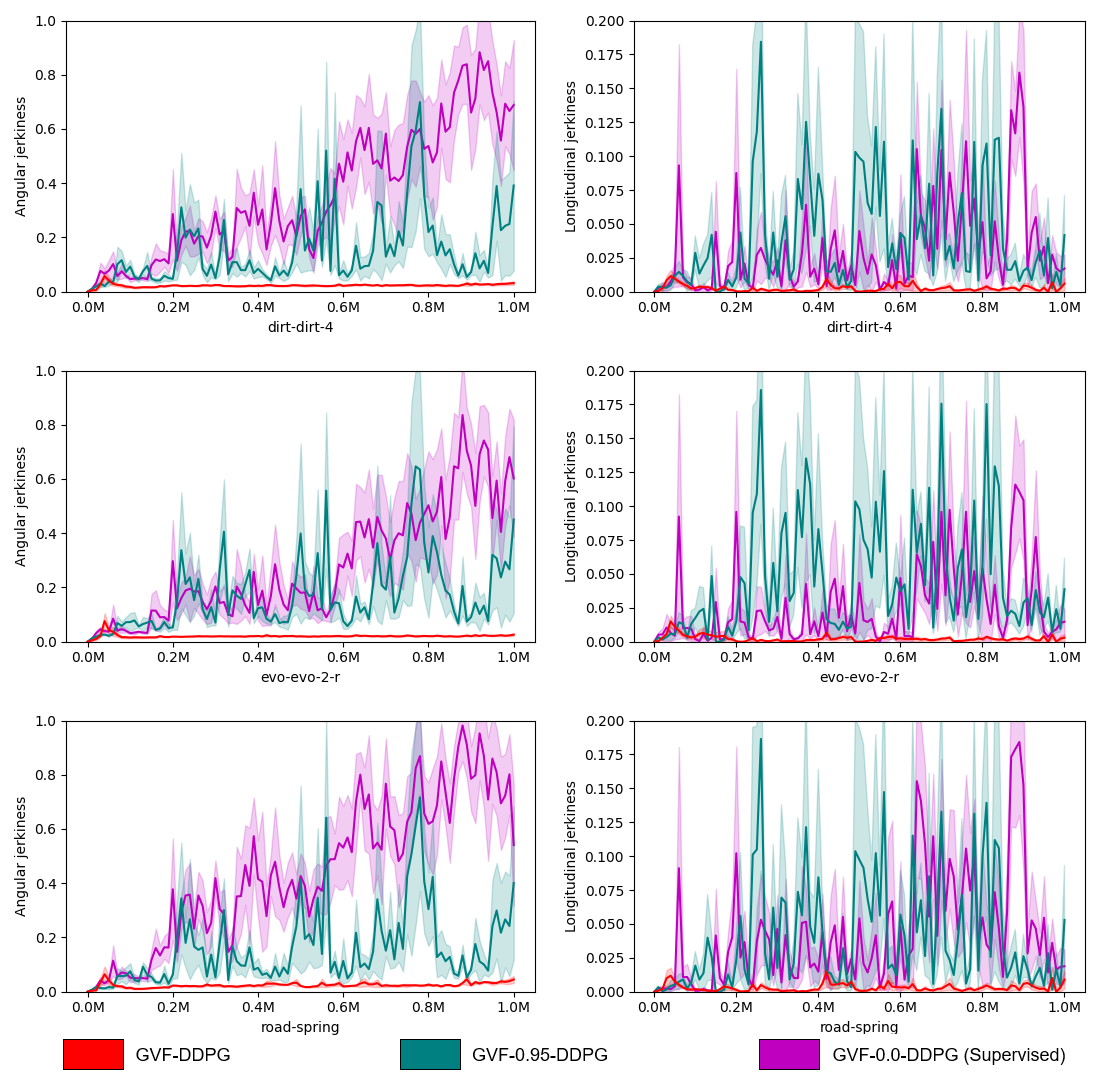}
	\caption{Ablation study of GVF-DDPG (our method) of jerkiness (lower is better) over different time scale selections. We use angular and longitudinal jerkiness to evaluate the smoothness of the learned policy. The jerkiness is evaluated every 1000 steps during training for dirt-dirt-4, evo-evo-2 and road-spring which were not part of the training set. Results show our proposed multi-time-scale predictions achieves the best performance.}
	\label{fig_jerkiness}
\end{figure}

The learned agents were trained on 85\% of 40 tracks available in TORCS.
The rest of the tracks were used for testing (6 in total) to measure the generalization performance of the policies.
Results are repeated over 5 runs for each method.
Only three of the tracks were successfully completed by at least one learned agent and those are reported here.
The reward in the TORCS environment is given by $r_t=0.0002 v_t (\cos{\beta_t} + |\alpha_t|)$ where $v_t$ is the speed of the vehicle in km/h, $\beta_t$ is the angle between the road direction and the vehicle direction, and $\alpha_t$ is the current lane centeredness.
The policies were evaluated on test roads at regular intervals during training as shown in Figures \ref{fig_summary_scores} and \ref{fig_jerkiness}.

The GVF-0.0-DDPG and GVF-0.95-DDPG variations initially learned very good solutions but then diverged indicating that one prediction may not be enough to control both steering angle and vehicle speed.
Despite an unfair advantage provided by DDPG-LowDim with the inclusion of lane centeredness and road angle in the observation vector, GVF-DDPG still outperforms both variants of DDPG on many of the test roads.
DDPG-Image was challenging to tune and train due to instability in learning; however, the counterfactual predictions in GVF-DDPG stabilized training for more consistent learning even though they were being learned simultaneously.
Only GVF-DDPG with multiple time scale predictions is able to achieve extraordinarily smooth control.

\section{Conclusions}
We present a new approach to learning to drive through a two step process: (1) learn a limited number of counterfactual predictions about future lane centeredness and road angle under a known policy, and (2) learn an RL policy using the counterfactual predictions as a representation of state.
Our novel approach is safe and practical because it learns from real-world driving data without online exploration where the behavior distribution of the driving data is unknown.
An experimental investigation into the impact of predictive representations on learning good driving policies shows that they generalize well to new roads, damaged lane markings and even distracting lane markings.
We find that our approach improves the performance, smoothness and robustness of the driving decisions from images.
We conclude that counterfactual predictions at different time scales is crucial to achieve a good driving policy.
To the best of our knowledge, this is the first practical demonstration of deep RL applied to autonomous driving on a real vehicle using only real-world data without any online exploration.

Our approach has the potential to be scaled with large volumes of data captured by human drivers of all skill levels; however, more work is needed to understand how well this approach will scale.
In addition, a general framework of learning the right counterfactual predictions for real-world problems is needed where online interaction is prohibitively expensive.

\bibliographystyle{IEEEtran}
\bibliography{IEEEabrv,references}

\clearpage
\appendix
\section{Appendix}

\subsection{Jackal Robot Experiments}
The Jackal robot is equipped with a 5MP camera using a wide angle lens, an IMU sensor and an indoor Hokuyo UTM-30LX LIDAR sensor with a $270^{\circ}$ scanning range and 0.1 to 10 meters scanning distance.
The objective is to drive the robot using a camera in the center of a lane marked with tape using only data collected in the real-world.
Training directly from real world experience requires addressing multiple issues such as minimizing wear and tear on the robot, and the need of human supervision during training in order to prevent or resolve robot crashes and recharge the battery.

There are two learned controllers, called GVF-BCQ and E2E-BCQ respectively, and one classical baseline called MPC (model predictive control).
The learned controllers output a steering angle $a_t^{steer}$ and target speed $a_t^{speed}$ based on the image taken by the camera in order drive centered in a closed loop road outlined with tape on a carpeted floor.
The MPC outputs a steering angle $a_t^{steer}$ and target speed $a_t^{speed}$ based on localization of the robot on a prior constructed map of the environment used to follow a sequence of waypoints supplied to the robot beforehand.
Localization, map and waypoints are needed to train the GVF-BCQ controller; however, this information is no longer used during testing.
GVF-BCQ tolerates noisy, low-accuracy localization methods that might otherwise be not accurate enough for smooth control with MPC or other methods.

\subsubsection{Training and testing environment}

The environment used for collecting data and evaluating the agents included two types of carpet floor - each with a different amount of friction.
The evaluation roads were done on one carpet only which was included in only about 20\% of the training data; the rest of the training data was on another type of carpet flooring to provide a generalization challenge for the learned controllers.
The friction was quite high on both carpets and it caused the agent to shake the camera violently while turning since the robot employs differential steering; tape on the wheels helped reduce the friction a bit.
Nevertheless, localization techniques using wheel odometry was deemed unsuitable and LIDAR-based localization was used instead.
LIDAR localization was not highly accurate but was sufficient; our tests showed that it
was repeatable to within roughly 5 centimeters which is deviation of upto about 13\% from the center of the road.

Nine closed loop roads were created by placing left and right lanes markings on the carpeted floor separated to form a consistent lane width of roughly 76 centimeters for all the roads; some error in lane width measurements were tolerated when creating our roads.
6 of the roads were selected for collecting data to train our learned agents.
Data was collected in both directions.
The remaining 3 roads were reserved for testing the performance of the policies.
In addition, each test road included damaged variants for a total of 6 test roads.

The poses and orientations of a sequence of waypoints were established to denote the center of lane which is the desired path of the agent; this was used to train our agents on the training roads and evaluate them on the test roads.
The center waypoints were collected by an expert manually and carefully navigating the Jackal robot on the road in the approximate center of lane; while this was inaccurate, it was not found to harm learning since the GVF-BCQ approach was able to generalize and learn features to drive the vehicle centered in the lane.
The LIDAR-based localization produced poses periodically to form the center waypoints; these were cleaned up by removing overlapping waypoints to form a closed loop path.
However, it did not produce poses at a consistent sampling frequency and thus a linear interpolation method was used to fill in the gaps and provide localization and center waypoint information at every time step.
The purpose of the center waypoints was to compute the road angle and lane centeredness of the robot in the road at any given time which is needed to train the GVF predictions and evaluate all of our controllers.

\begin{figure}[t]
	\centering
	\includegraphics[width=8cm]{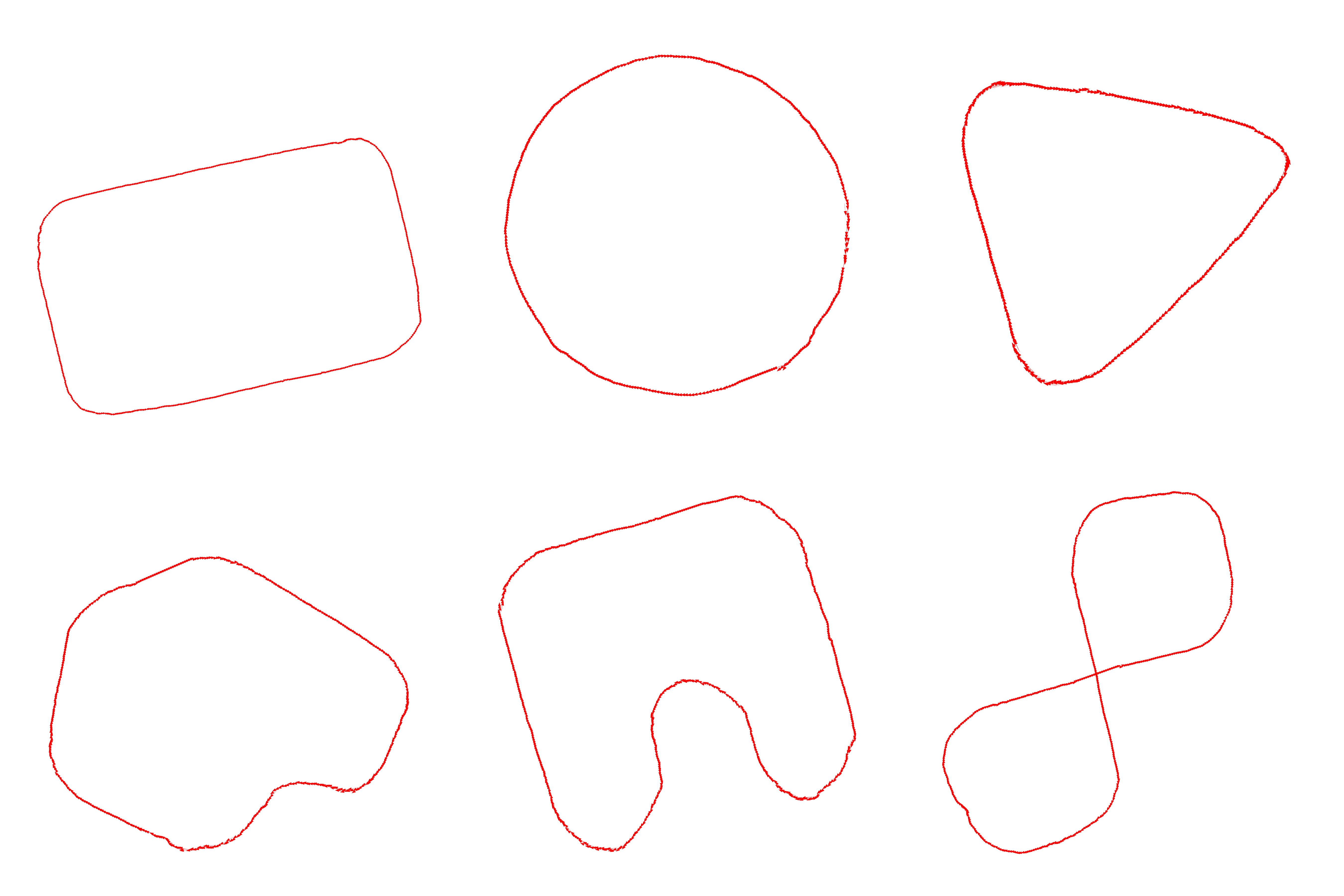}
	\caption{Six roads for training.  The second row shows more complex road structure.  The rectangle road has rectangular edges at all corners.}
	\label{fig_jackal_train_tracks}
\end{figure}

\begin{figure}[t]
	\centering
	\includegraphics[width=8cm]{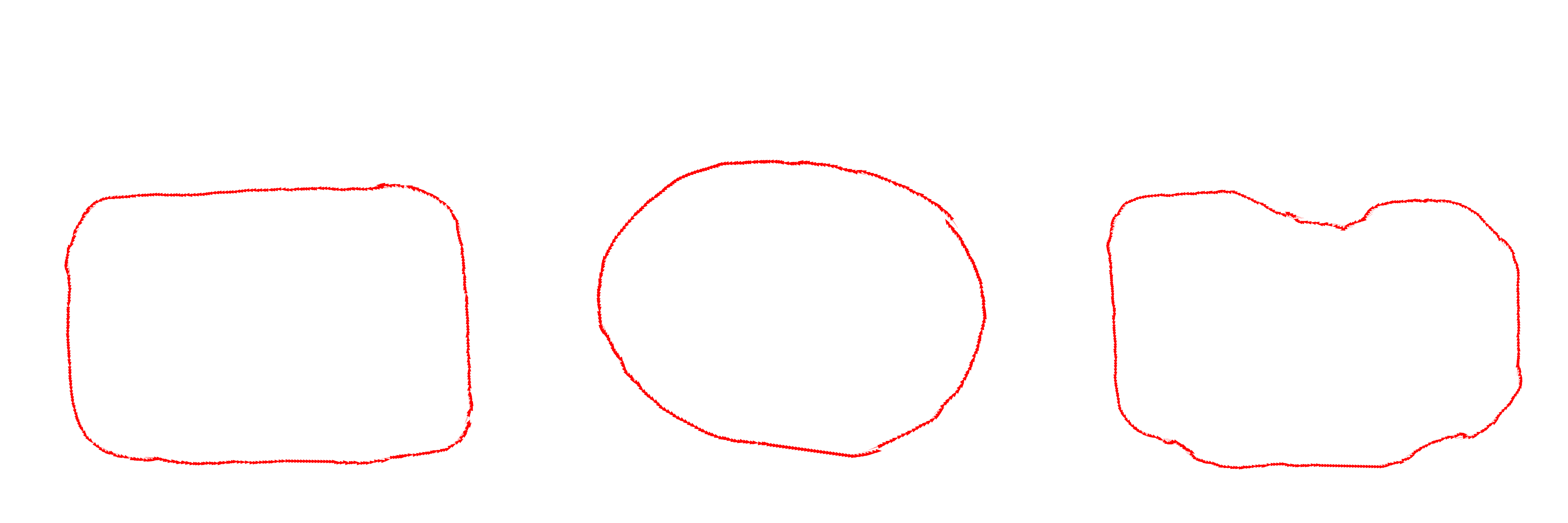}
	\caption{Three roads for testing.  Left to right: (1) rectangle with rounded corners; (2) oval; (3) complex.}
	\label{fig_jackal_test_tracks}
\end{figure}

\begin{figure}[t]
	\centering
	\includegraphics[width=8cm]{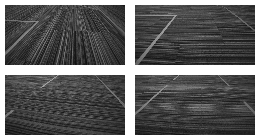}
	\caption{Input images of normal lane markings (top row) and damaged lane markers (bottom row).}
	\label{fig_jackal_train_tracks_camera}
\end{figure}

The center waypoints for the training and testing roads are depicted in Figure \ref{fig_jackal_train_tracks} and Figure \ref{fig_jackal_test_tracks}, respectively.
The first three training roads were simple geometric shapes while the other three were a bit more complex.
The first test road was the most similar to the training data where the outer edge of the four corners were rounded.
The second test road was an oval shape to evaluate how well the agent maintained a turn that, unlike the circle training road, requires the steering angle to be modulated rather than remain constant.
The third test road was a complex shape with multiple sudden turns that was very different from any of the roads in the training data set.
This tests generalization to new roads and confirms that the agent is not memorizing an action sequence to remain centered in the path.
All methods were evaluated at 0.25 m/s and 0.4 m/s maximum speeds and clock-wise (CW) and counter clock-wise (CCW) directions.
In order to test robustness, all three test roads were altered by degrading or damaging the lane markings.
The complex test road also included distracting markings as shown in Figure \ref{fig_jackal_overview}.
An example of an image received from the robot with and without damage to the lane markers is shown in Figure \ref{fig_jackal_train_tracks_camera}.

\subsubsection{Data collection}
Both learned agents -- GVF-BCQ and E2E-BCQ -- were trained with batch RL \cite{fujimoto2018off} where the GVF-BCQ method learned a predictive representation.
Rosbags were collected with the Jackal robot each containing around 30 minutes to 2 hours of data for a total of 40 hours of data containing approximately 1.5 million images.
The camera image, localization pose, IMU measurement, and action taken by the controller was recorded at each time step.
The Jackal robot was controlled using a random walk controller that was confined to the road area to provide sufficient and safe exploration.
The map was created with Hector SLAM \cite{KohlbrecherMeyerStrykKlingaufFlexibleSlamSystem2011} and the localization produced by Adaptive Monte-Carlo Localization \cite{thrun2002probabilistic}.

The random walk controller was based on a pure pursuit controller, where action taken at each time step is defined by
\begin{equation}
\begin{split}
a_t^{steer} & = \text{clip}(\text{angle}(p_t, p^*_{k(t)}) - \theta^z_t, -\pi / 2, \pi / 2) \\
a_t^{speed} & = \text{clip}(v^*_{k(t)}, 0.2, 0.5)
\end{split}
\label{eq_jackal_pure_pursuit_actions}
\end{equation}
where $\theta^z_t$ and $p_t$ were the yaw angle and the 2-dimensional position of the robot in the real world (obtained from localization), $p^*_{k(t)}$ and $v^*_{k(t)}$ were the target position and linear velocity at time $t$, clip($x$, MinVal, MaxVal) is the clip function that operates in scalar and vector element-wise, and angle$(p_t, p^*_{k(t)})$ is the function that returns the yaw angle in the real world of a vector that points from $p_t$ to $p^*_{k(t)}$.
The target position $p^*_{k(t)}$ and linear velocity $v^*_{k(t)}$ were encapsulated in the next target pose at index $k(t)$ in the sequence of target poses:
\begin{equation}
\begin{split}
k(1) & = 1 \\
k(t+1) & = 
	\begin{cases}
k(t) + 1 \text{ if } ||p_t-p^*_{k(t)}||_2 < 0.025 \\
k(t) \text{ otherwise }
	\end{cases}
\end{split}
\label{eq_jackal_waypoint_update}
\end{equation}
Thus, the robot advanced to the next target position and linear velocity in the target pose sequence once it arrived within 2.5 centimeters of the current target position.
In order to provide efficient and safe exploration that can be confined to the road area, the target position $p^*_j$ was based on the position $\tilde{p}_j$ of the center waypoints collected earlier with some noise added to provide the necessary exploration to learn the predictions and policy:
\begin{equation}
\begin{split}
p^*_{j} & = \tilde{p}_{j \% N} + \varepsilon^p_j \\
v^*_{j} & = 
	\begin{cases}
		v^*_{j-1} + \varepsilon^v_j \text{ if } j > 1 \\
		0.35 \text{ if } j = 1
	\end{cases}
\end{split}
\label{eq_jackal_target_noise}
\end{equation}
where $N$ is the number of points that define the center waypoints of the closed loop road.
$\varepsilon^p_j$ and $\varepsilon^v_j$ were the noises added at each time step:
\begin{equation}
\begin{split}
	\varepsilon^p_j & = 
		\begin{cases}
			\text{clip}(\varepsilon^p_{j-1} + \mathcal{N}(0, 0.02 * \mathbbm{1}), -0.3, 0.3) \text{ if } j > 1 \\
			[0, 0]^\intercal \text{ if } j = 1
		\end{cases} \\
	\varepsilon^v_j & = \mathcal{N}(0, 0.02)
\end{split}
\label{eq_jackal_exploration_noise}
\end{equation}
The noises for the poses were clipped so that the robot would not explore too far outside the road area.

The rosbags were processed to synchronize the sensor data streams at a fixed sample frequency of 10Hz and compute the lane centeredness $\alpha_t$, road angle $\beta_t$, and speed $v_t$ of the robot at each time step:
\begin{equation}
\begin{split}
	\nu_t & = \text{knn}(p_t, S_t) \\
	\alpha_t & = \text{clip}(\frac{||p_t - \nu_t||_2}{H}, -1.0, 1.0) \\
	\beta_t & = \text{clip}(\text{angle}(p_t, \nu_t) - \theta_t^z, -\pi/2, \pi/2) \\
\end{split}
\end{equation}
where $\text{knn}(x, S)$ returns $\nu$ as the closest point to $x$ in $S$ using k-nearest neighbor and $H=38$ centimeters as the half lane width.
$S_t$ is a pruned set of center waypoints where $S_t = \{\tilde{p}_t\}$ for all roads, except for the figure 8 road in the lower right of Figure \ref{fig_jackal_train_tracks} where $S_t$ was based on a sliding window to prevent issues with k-nn at the intersection:
\begin{equation}
\begin{split}
	S_t = 
		\begin{cases}
			\{\tilde{p}_t\} \text{ if } j=1 \\
			\{\tilde{p}_{t=I_{t-1}-w \rightarrow I_{t-1}+w}\} \text{ if } j > 1
		\end{cases}
\end{split}
\end{equation}
where $I_{j-1}$ is the index of $\nu_{t-1}$ in $S_{t-1}$ at the previous time step and $w=10$ is the size of the sliding window.
Negative indices are wrapped to the beginning of the waypoint list.
The speed was estimated using the change in position over a single time step which was quite noisy but more reliable than speed returned from the robot's odometry.
Due to computation constraints on the robot, the localization messages were output at less than 10Hz; thus, a linear interpolation was used to fill in missing poses and orientations in the data and synchronize the data streams.

Images from camera with original size $1920 \times 1080$ were cropped to $960 \times 540$ region in the center and then additionally top-cropped by 60.
The images were then downsampled by a factor of 8 in both spatial dimensions to give a final size of $120 \times 60$ and then converted to gray scale.
To improve generalization in deep learning and balance the left-right biases in the data, augmented data was created with horizontally flipped images along with the corresponding signs of the lane centeredness $\alpha_t$, road angle $\beta_t$, and steering action $a_t^{steer}$) flipped.

\subsubsection{GVF-BCQ Training}
The predictive neural network used was trained using the offline version of the predictive learning algorithm \ref{alg_gvf_offpolicy_train_without_mu_offline}.
The transitions in the data were loaded into a replay buffer in the same order that the transitions were observed in the rosbag where mini-batches of size 128 where sampled from the growing replay buffer and used to update the GVFs.
The GVFs were updated for 5 million steps followed by BCQ for an additional 5 million steps for a total of 10 million steps.
The order that the rosbags were loaded into the replay buffer was randomized.
The replay buffer had a maximum capacity of 0.5 million samples; once the replay buffer was filled, the oldest samples were removed.
Training began once the replay buffer reached 0.1 million samples.
While an alternative approach would have been to sample mini-batches from the entire data set from the beginning, our approach was found to be effective and required minimal changes to the data loader of the online version of the algorithm.

\begin{figure}[t]
	\centering
	\includegraphics[width=6cm]{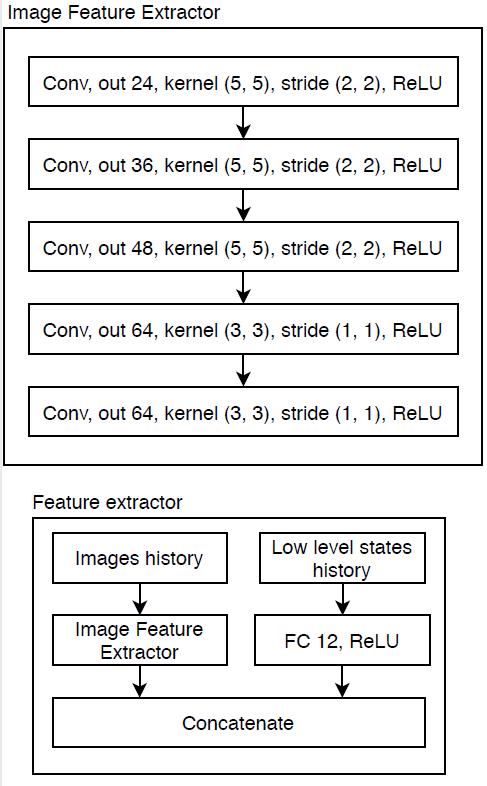}
	\caption{Model of the feature extractor of the robot's state concatenates features from the image with low-dimensional state information from the robot like speed and last action.}
	\label{fig_feature_extractor_model}
\end{figure}
\begin{figure}[t]
	\centering
	\begin{subfigure}[b]{4cm}
		\includegraphics[width=2cm]{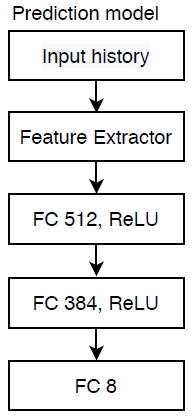}
		\caption{Prediction Model}
		\label{fig_prediction_model}
	\end{subfigure}
	\begin{subfigure}[b]{4cm}
		\includegraphics[width=2cm]{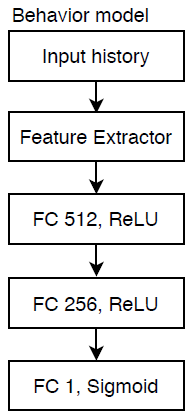}
		\caption{Behavior Model}
		\label{fig_behavior_model}
	\end{subfigure}
	\caption{Neural network models for (a) Prediction Model that produces $\bm{\psi}(s)$, and (b) Behavior Model that estimates $\mu(a|s)$}
	\label{fig_pred_behav_models}
\end{figure}
\begin{figure}[t]
	\centering
	\includegraphics[width=4cm]{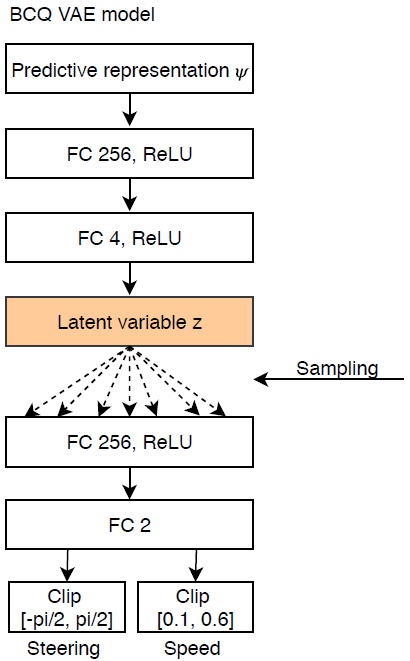}
	\caption{Model of the GVF-BVQ variational auto-encoder (VAE).}
	\label{fig_gvfbcq_vae_model}
\end{figure}
\begin{figure}[t]
	\centering
	\begin{subfigure}[b]{4cm}
		\includegraphics[width=4cm]{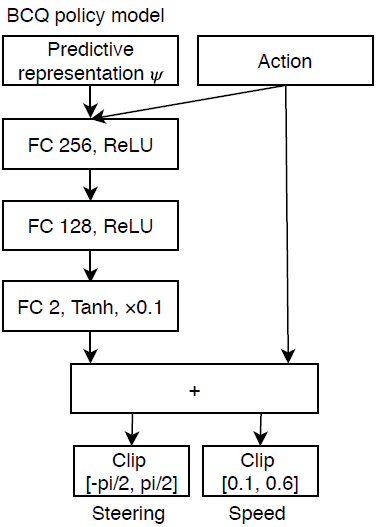}
		\caption{Actor Model}
		\label{fig_gvfbcq_actor_model}
	\end{subfigure}
	\begin{subfigure}[b]{4cm}
		\includegraphics[width=4cm]{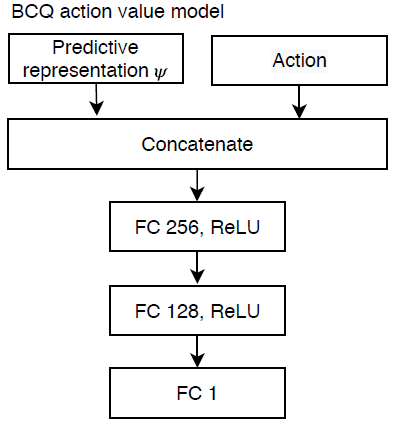}
		\caption{Critic Model}
		\label{fig_gvfbcq_critic_model}
	\end{subfigure}
	\caption{Neural network models for GVF-BCQ (a) Actor, and (b) Critic}
	\label{fig_gvfbcq_actor_critic_models}
\end{figure}

The feature extractor model of the robot state is depicted in Figure \ref{fig_feature_extractor_model}.
Estimating the behavior distribution for the GVF predictions was done with $\eta(a|s)$ as a uniform distribution defined on the interval $[-\pi/2,\pi/2]$ for the steering action and a uniform distribution defined on the interval $[0, 1]$ for the target speed action.
The neural network model used to learn the GVFs predictions that produce $\phi$ are depicted in Figure \ref{fig_prediction_model}.
The model used to estimate the behavior distribution $\mu(a|s)$ is shown in Figure \ref{fig_behavior_model}.
The BCQ network models that are used to learn the policy of the agent were all relatively small fully connected networks with hidden layer of size 256 as shown in Figures \ref{fig_gvfbcq_vae_model}, \ref{fig_gvfbcq_actor_model}, and \ref{fig_gvfbcq_critic_model}.

\begin{figure*}[t]
	\centering
	\includegraphics[width=16cm]{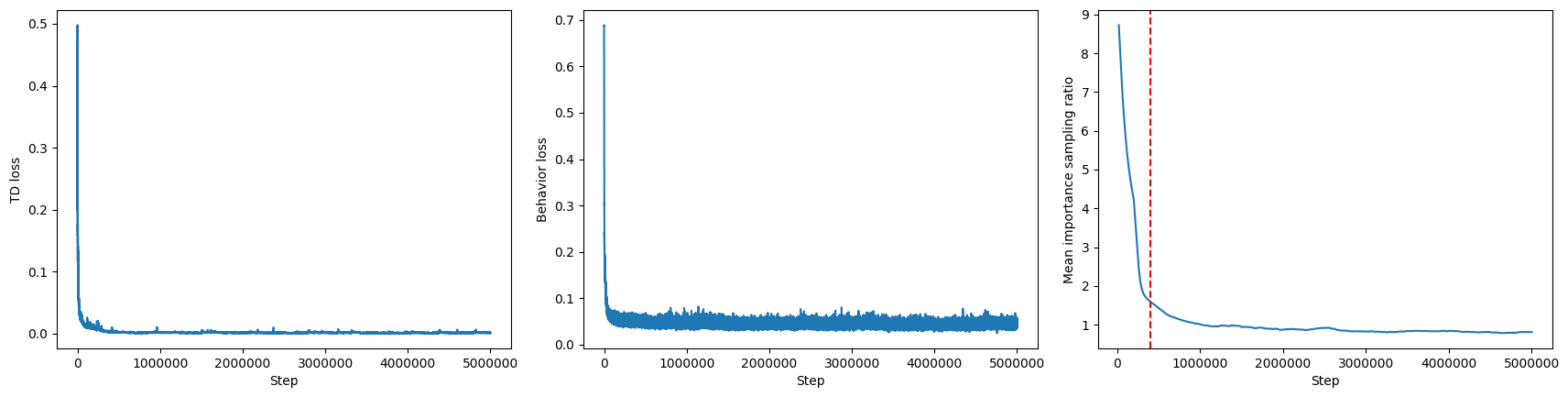}
	\caption{TD loss, behavior loss and mean importance sampling ratio in the buffer over training steps. Red vertical dash line is the point when the buffer is full.}
	\label{fig_jackal+is_ratio}
\end{figure*}

The predictive representation $\psi$ is a vector of length 11 consisting of $\phi$ (vector of predictions of size 8), the last steering action, the last target speed action and the current robot speed.
The latent vector dimension was 4 which was a Normal distribution parameterised by mean and log standard deviation. 
All networks used ReLU activation for the hidden layers and linear activation for the outputs.
The action output from the actor and VAE were clipped to $[-\pi/2, \pi/2]$ for steering and $[0.1, 0.6]$ for the target speed.
The weight of the KL divergence loss used in BCQ was 0.5.
The learning rate was $10^{-4}$ for both GVF and BCQ model training.
Figure \ref{fig_jackal+is_ratio} shows the training curves for the predictive representation (GVFs) including the temporal-difference loss, behavior model loss and mean importance sampling ratio.

\subsubsection{End-to-end BCQ Baseline Training}
\begin{figure}[t]
	\centering
	\begin{subfigure}[b]{3cm}
		\includegraphics[width=3cm]{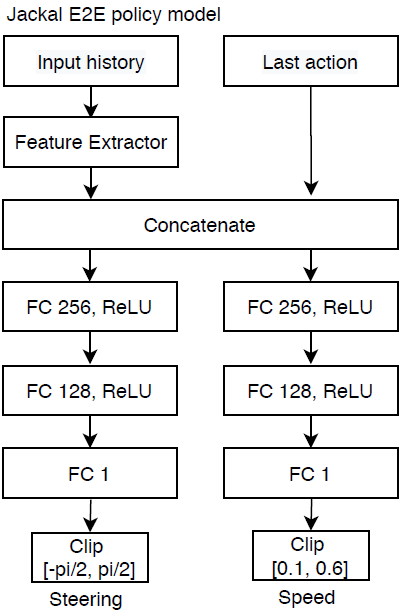}
		\caption{Actor Model}
		\label{fig_e2ebcq_actor_model}
	\end{subfigure}
	\begin{subfigure}[b]{4cm}
		\includegraphics[width=4cm]{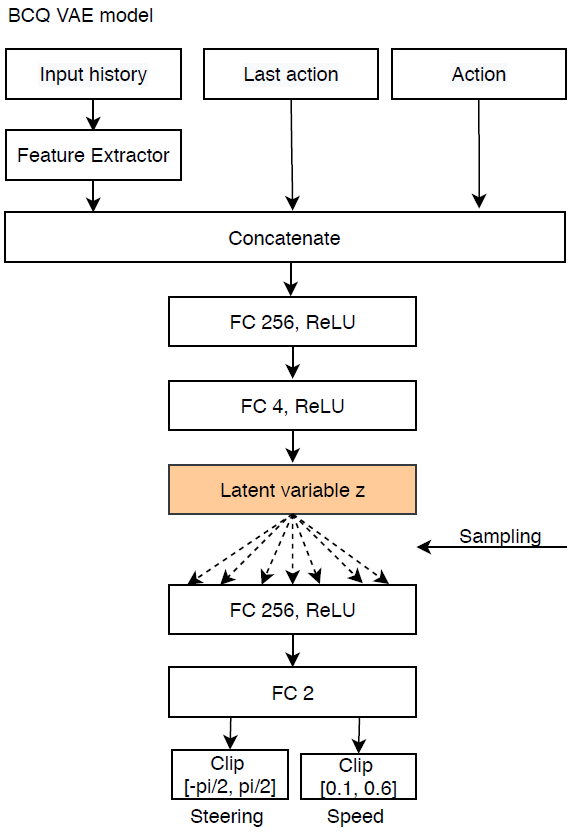}
    	\caption{VAE Model}
    	\label{fig_e2ebcq_vae_model}
	\end{subfigure}
	\caption{Neural network models for E2E-BCQ (a) Actor, and (b) VAE.  The Critic model is the same as DDPG in Figure \ref{fig_ddpg_critic_model}}
	\label{fig_e2ebcq_actor_critic_models}
\end{figure}

Nearly the same training setup used for GVF-BCQ was also applied to the E2E-BCQ method.
The hyperparameters, training settings, activation functions for the output and action clipping are exactly the same as GVF-BCQ unless noted otherwise.
All the networks in E2E-BCQ including the VAE, actor and critic shared the same feature extractor as the GVF-BCQ shown in Figure \ref{fig_feature_extractor_model}.
The neural network models are given in Figures \ref{fig_e2ebcq_vae_model}, and \ref{fig_e2ebcq_actor_model}.

For a fair comparison, E2E-BCQ was trained for 10 millions update steps - the same number as GVF-BCQ method which divides the update budget evenly where 5 million updates are applied to the GVF predictions and 5 million updates are applied to BCQ networks.
The agent was then tested on the rectangle test road and it was found that E2E-BCQ performed very poorly. 
The agent was too slow reaching an average speed of about 0.18 m/s whereas the GVF-BCQ method was able to reach double that speed.
In addition, E2E-BCQ steered very poorly and was often not centered in the lane; unlike the GVF-BCQ method which was observed to be quite robust, the E2E-BCQ method sometimes drove out of the lane where an emergency stop was needed to prevent collision.
For this reason, E2E-BCQ was only compared to GVF-BCQ on the rectangle test road; and we focused on comparisons against the MPC controller which was more robust than E2E-BCQ.
A detailed evaluation of E2E-BCQ is shown in Figure \ref{fig_jackal_image_bcq} and Table \ref{table_summary_jackal_result_gvf_vs_e2e}.

\begin{figure*}[t]
	\centering
	\includegraphics[width=16cm]{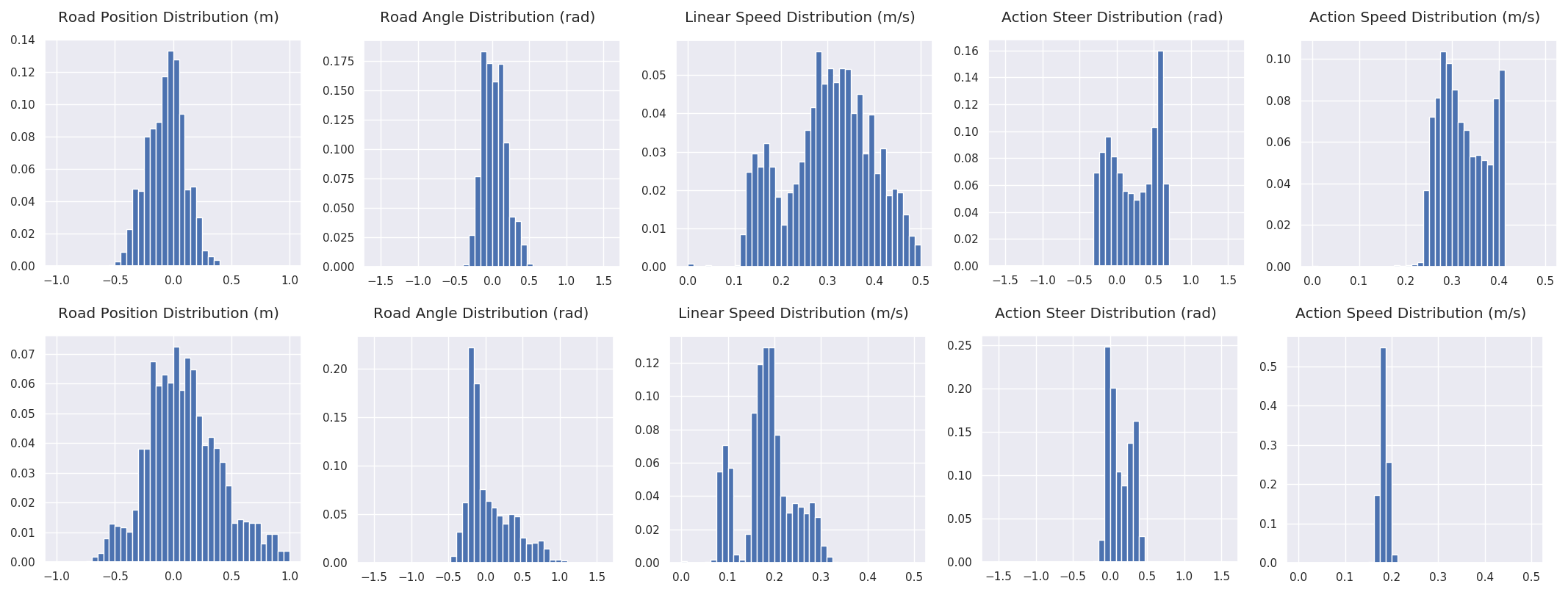}
	\caption{Distribution of lane centeredness, road angle, speed and action distribution of GVF-BCQ and E2E-BCQ on the rectangle test track at 0.4 speed, counterclockwise direction.}
	\label{fig_jackal_gvf_vs_e2e}
\end{figure*}

\subsubsection{MPC Baseline}
An MPC baseline using standard ros nodes available for the Jackal robot were used for controlling the Jackal robot.
The baseline was tuned for 0.4 m/s; however, it was challenging to achieve good performance due to limited computation power for the look ahead, noisy localization with LIDAR scan matching and inaccurate modeling of the steering characteristics on carpet floors.
The best performance was achieved for 0.25 m/s but significant oscillation was observed for 0.4 m/s that was very challenging to completely eliminate.
The center waypoints provided as input to the MPC controller were processed so that the minimum distance between two consecutive waypoints was 2.5cm; the waypoints were then downsampled by a factor of 16 in order to increase their separation and increase the look ahead distance; different downsampling factors was tested but oscillation was never completely eliminated.
The MPC had an optimization window of 5 steps into the future; this was limited by computation available on the Jackal robot's on-board computer.
This look ahead ensured the MPC was far enough into the future for real-time control.

\subsubsection{Test Results}

This section provides a detailed comparison of GVF-BCQ and the baselines at different speeds and directions.
For both GVF-BCQ and E2E-BCQ methods, the actor produced the steering and desired speed and thus the agent was able to modulate its own speed and slow down as necessary in advance of sharp turns.
The speed command was clipped at the maximum target speed.
The controllers started at the same position and heading angle and they were allowed to run for exactly 300 seconds.
The agents were evaluated based on the following criteria:

\begin{itemize}
\item Reward per second: $\frac{1}{N} \sum_{t=1}^N r_t$
\item Average speed: $\frac{1}{N} \sum_{t=1}^N v_t$
\item Average absolute lane centeredness: $\frac{1}{N} \sum_{t=1}^N |\alpha_t|$
\item Average absolute road angle: $\frac{1}{N} \sum_{t=1}^N |\beta_t|$
\item Near out of lane\footnote{ratio of time steps where the agent's absolute lane centeredness is greater than 0.75}:  $\frac{1}{N} \sum_{t=1}^N \mathbbm{1}_{|\alpha_t| > 0.75}$\footnote{Where $\mathbbm{1}$ is the indicator function.}.
\item First Order Jerk\footnote{First order jerk is the absolute change of action taken by the agent in one time step.  Lower jerk scores are better.  Both steering and speed actions considered separately.}: $\frac{1}{N-1} \sum_{t=1}^{N-1} |a_{t+1}-a_t|$
\item Second Order jerk\footnote{Second order jerk is the absolute change of the first order jerk in one time step.  Lower jerk scores are better.  Both steering and speed actions considered separately.}: $\frac{1}{N-2} \sum_{t=1}^{N-2} |(a_{t+2}-a_{t+1})-(a_{t+1}-a_t)|$
\end{itemize}

\begin{table*}[t]
\centering
\caption{Comparison of GVF-BCQ and E2E-BCQ on the Rectangle test road at 0.4 m/s}
\label{table_jackal_result_gvf_vs_e2e1}
\begin{tabular}{|l|l|l|l|l|l|l|} 
\hline
    & Experiment & \multicolumn{1}{c|}{\begin{tabular}[c]{@{}c@{}}Reward\\ per \\ second \textuparrow \end{tabular}} & \multicolumn{1}{c|}{\begin{tabular}[c]{@{}c@{}}Average\\ speed \textuparrow \end{tabular}} & \multicolumn{1}{c|}{\begin{tabular}[c]{@{}c@{}}Average \\ off-center\\ (normalized) \\ \textdownarrow \end{tabular}} & \multicolumn{1}{c|}{\begin{tabular}[c]{@{}c@{}}Average \\ off-angle \\ \textdownarrow \end{tabular}} & \multicolumn{1}{c|}{\begin{tabular}[c]{@{}c@{}}Out of \\ lane \textdownarrow \end{tabular}}  \\ 
\hline
\multirow{4}{*}{\begin{tabular}[c]{@{}l@{}}Rectangle\\ shape \end{tabular}}
    & GVF-BCQ-0.4-CCW& \textbf{2.6835 }& \textbf{0.3205 }& \textbf{0.1345 }& \textbf{0.1315 }& \textbf{0.0\% }\\
    & E2E-BCQ-0.4-CCW& 1.2578& 0.1816& 0.2558& 0.2414& 3.76\%\\ 
\cline{2-7}
    & GVF-BCQ-0.4-CW & \textbf{2.2915} & \textbf{0.3140} & \textbf{0.2217} & \textbf{0.1586} & \textbf{0.0\%} \\
    & E2E-BCQ-0.4-CW & -0.1302& 0.1710& 0.9927& 0.3034& 54.18\%\\
\hline
\end{tabular}
\end{table*}

\begin{table*}[t]
\centering
\caption{Comparison of GVF-BCQ and E2E-BCQ jerk levels on the Rectangle test road at 0.4 m/s}
\label{table_jackal_result_gvf_vs_e2e2}
\begin{tabular}{|l|l|l|l|l|l|} 
\hline
    & Experiment & \multicolumn{1}{c|}{\begin{tabular}[c]{@{}c@{}}First \\ order\\ speed\\ jerk \textdownarrow \end{tabular}} & \multicolumn{1}{c|}{\begin{tabular}[c]{@{}c@{}}Second \\ order\\ speed \\ jerk \textdownarrow \end{tabular}} & \multicolumn{1}{c|}{\begin{tabular}[c]{@{}c@{}}First\\ order\\ steering\\ jerk \textdownarrow \end{tabular}} & \multicolumn{1}{c|}{\begin{tabular}[c]{@{}c@{}}Second\\ order\\ steering \\ jerk \textdownarrow \end{tabular}}  \\ 
\hline
\multirow{4}{*}{\begin{tabular}[c]{@{}l@{}}Rectangle\\ shape \end{tabular}}
    & GVF-BCQ-0.4-CCW& \textbf{0.0356 }& \textbf{0.2532 } & \textbf{0.2251 } & \textbf{1.3403 } \\
    & E2E-BCQ-0.4-CCW& 0.0154  & 0.2266 & 0.1109 & 1.4240 \\ 
\cline{2-6}
    & GVF-BCQ-0.4-CW & \textbf{0.0311}& \textbf{0.2149}  & \textbf{0.1995}  & \textbf{1.1850}  \\
    & E2E-BCQ-0.4-CW & 0.0148  & 0.1937 & 0.1174 & 1.3514 \\
\hline
\end{tabular}
\end{table*}

A comparison of GVF-BCQ and E2E-BCQ is given in Tables \ref{table_jackal_result_gvf_vs_e2e1} and \ref{table_jackal_result_gvf_vs_e2e2}.
Experiments are named according to the method used, the selected target speed and the direction of the road loop (i.e. counter-clock-wise versus clock-wise).
For example, GVF-BCQ-0.4-CCW points to the test of the GVF-BCQ controller with 0.4 m/s target speed in the counter-clock-wise direction.

\begin{table*}[t]
\centering
\caption{Evaluation of the robustness of GVF-BCQ method on damaged lane markings on all the test roads}
\label{table_jackal_result_N_vs_D1}
\begin{tabular}{|l|l|l|l|l|l|l|} 
\hline
    & Experiment & \multicolumn{1}{c|}{\begin{tabular}[c]{@{}c@{}}Reward\\ per \\ second \textuparrow \end{tabular}} & \multicolumn{1}{c|}{\begin{tabular}[c]{@{}c@{}}Average\\ speed \textuparrow \end{tabular}} & \multicolumn{1}{c|}{\begin{tabular}[c]{@{}c@{}}Average \\ off-center\\ (normalized) \\ \textdownarrow \end{tabular}} & \multicolumn{1}{c|}{\begin{tabular}[c]{@{}c@{}}Average \\ off-angle \\ \textdownarrow \end{tabular}} & \multicolumn{1}{c|}{\begin{tabular}[c]{@{}c@{}}Out of\\ lane \textdownarrow \end{tabular}}  \\ 
\hline
\multirow{2}{*}{\begin{tabular}[c]{@{}l@{}}Rectangle\\ shape \end{tabular}}
    & GVF-BCQ-0.4-CCW& 2.6835& 0.3205& \textbf{0.1345 }& \textbf{0.1315 }& 0.0\%\\
    & GVF-BCQ-0.4-CCW-D & \textbf{2.7407 }& \textbf{0.3261 }& 0.1358& 0.1351& 0.0\%\\ 
\hline
\multirow{2}{*}{\begin{tabular}[c]{@{}l@{}}Oval\\ shape \end{tabular}}
    & GVF-BCQ-0.4-CCW& \textbf{2.4046 }& \textbf{0.3501 }& \textbf{0.2754 }& 0.2125& \textbf{1.45\% }\\
    & GVF-BCQ-0.4-CCW-D & 2.0728& 0.3279& 0.3285& \textbf{0.2089 }& 7.19\%\\ 
\hline
\multirow{2}{*}{\begin{tabular}[c]{@{}l@{}} Complex\\ shape \end{tabular}}
    & GVF-BCQ-0.4-CCW& \textbf{2.3501 }& 0.3129& \textbf{0.2221 }& \textbf{0.1817 }& \textbf{0.0\% }\\
    & GVF-BCQ-0.4-CCW-D & 2.1059& \textbf{0.3284 }& 0.3125& 0.2365& 9.42\%\\
\hline
\end{tabular}
\end{table*}

\begin{table*}[t]
\centering
\caption{Evaluation of the jerk of GVF-BCQ method on damaged lane markings on all the test roads}
\label{table_jackal_result_N_vs_D2}
\begin{tabular}{|l|l|l|l|l|l|} 
\hline
    & Experiment & \multicolumn{1}{c|}{\begin{tabular}[c]{@{}c@{}}First \\ order\\ speed\\ jerk \textdownarrow \end{tabular}} & \multicolumn{1}{c|}{\begin{tabular}[c]{@{}c@{}}Second \\ order\\ speed \\ jerk \textdownarrow \end{tabular}} & \multicolumn{1}{c|}{\begin{tabular}[c]{@{}c@{}}First\\ order\\ steering\\ jerk \textdownarrow \end{tabular}} & \multicolumn{1}{c|}{\begin{tabular}[c]{@{}c@{}}Second\\ order\\ steering \\ jerk \textdownarrow \end{tabular}}  \\ 
\hline
\multirow{2}{*}{\begin{tabular}[c]{@{}l@{}}Rectangle\\ shape \end{tabular}}
    & GVF-BCQ-0.4-CCW& \textbf{0.0356 }& \textbf{0.2532 } & \textbf{0.2251 } & \textbf{1.3403 } \\
    & GVF-BCQ-0.4-CCW-D & 0.0383  & 0.2715 & 0.2303 & 1.4620 \\ 
\hline
\multirow{2}{*}{\begin{tabular}[c]{@{}l@{}}Oval\\ shape \end{tabular}}
    & GVF-BCQ-0.4-CCW& 0.0348  & \textbf{0.2423 } & 0.2191 & \textbf{1.4632 } \\
    & GVF-BCQ-0.4-CCW-D & \textbf{0.0334 }& 0.2953 & \textbf{0.2094 } & 1.6612 \\ 
\hline
\multirow{2}{*}{\begin{tabular}[c]{@{}l@{}} Complex\\ shape \end{tabular}}
    & GVF-BCQ-0.4-CCW& \textbf{0.0341 }& \textbf{0.2540 } & \textbf{0.2272 } & \textbf{1.5306 } \\
    & GVF-BCQ-0.4-CCW-D & 0.0437  & 0.3608 & 0.2897 & 2.0946 \\
\hline
\end{tabular}
\end{table*}

\begin{figure*}[t]
	\centering
	\includegraphics[width=16cm]{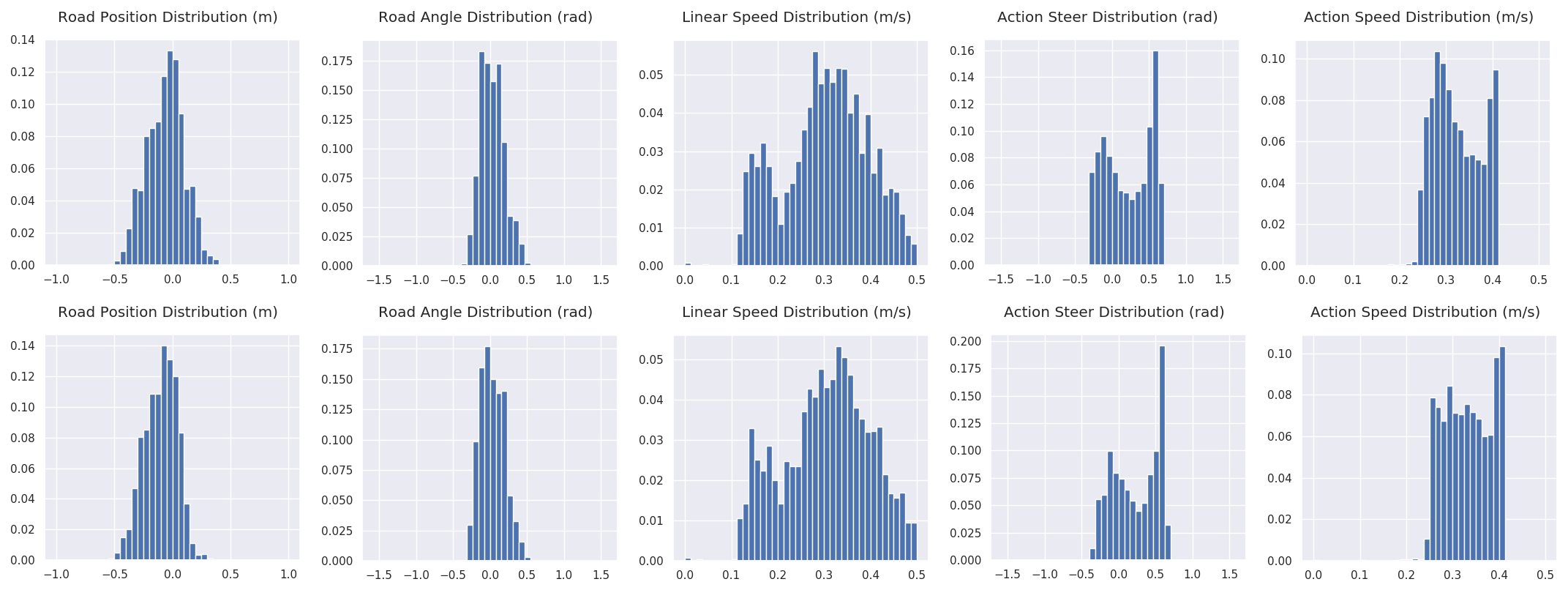}
	\caption{Distribution of lane centeredness, road angle, speed and action distribution on the rectangle test road. From top to bottom: GVF-BCQ-0.4, GVF-BCQ-0.4 with lane marking damage on the rectangle road. The similarities highlight the robustness of GVF-BCQ to the introduction of damaged lanes.}
	\label{fig_jackal_gvf_N_vs_D}
\end{figure*}

Finally, the GVF-BCQ method generalized well to damaged lane markings and distractions in the visual images as shown in the similar scores and similar distributions in Figure \ref{fig_jackal_gvf_N_vs_D} and Table \ref{table_jackal_result_N_vs_D1} and \ref{table_jackal_result_N_vs_D2}.
Experiments on roads with damaged lane markings are denoted with suffix -D.

\begin{table*}[t]
\centering
\caption{Comparison of GVF-BCQ method and MPC on all the test roads at different speeds and directions}
\label{table_jackal_result_gvf_vs_mpc1}
\begin{tabular}{|l|l|l|l|l|l|l|} 
\hline
    & Experiment & \multicolumn{1}{c|}{\begin{tabular}[c]{@{}c@{}}Reward\\ per \\ second \textuparrow \end{tabular}} & \multicolumn{1}{c|}{\begin{tabular}[c]{@{}c@{}}Average\\ speed \textuparrow \end{tabular}} & \multicolumn{1}{c|}{\begin{tabular}[c]{@{}c@{}}Average \\ off-center\\ (normalized) \\ \textdownarrow \end{tabular}} & \multicolumn{1}{c|}{\begin{tabular}[c]{@{}c@{}}Average \\ off-angle\\ \textdownarrow \end{tabular}} & \multicolumn{1}{c|}{\begin{tabular}[c]{@{}c@{}}Out of\\lane \textdownarrow \end{tabular}}  \\ 
\hline
\multirow{6}{*}{\begin{tabular}[c]{@{}l@{}}Rectangle\\ shape \end{tabular}}
    & GVF-BCQ-0.4-CCW & \textbf{2.6835} & 0.3205 & \textbf{0.1345} & \textbf{0.1315} & \textbf{0.0\%}\\
    & MPC-0.4-CCW & 0.9700 & \textbf{0.3833} & 0.5252 & 0.1943 & 20.42\%\\ 
\cline{2-7}
    & GVF-BCQ-0.4-CW & \textbf{2.2915} & 0.3140 & \textbf{0.2217} & \textbf{0.1586} & \textbf{0.0\%}\\
    & MPC-0.4-CW & 0.1282 & \textbf{0.3836} & 0.9086 & 0.1916 & 67.86\%\\ 
\cline{2-7}
    & GVF-BCQ-0.25-CCW & \textbf{2.1442} & \textbf{0.2467} & \textbf{0.1098} & \textbf{0.1181} & 0.0\%\\
    & MPC-0.25-CCW & 1.1971& 0.2412& 0.1218& 0.1308& 0.0\%\\ 
\hline
\multirow{6}{*}{\begin{tabular}[c]{@{}l@{}}Oval\\ shape \end{tabular}}
    & GVF-BCQ-0.4-CCW  & \textbf{2.4046} & 0.3501& \textbf{0.2754} & 0.2125& \textbf{1.45\%} \\
    & MPC-0.4-CCW  & 0.8928& \textbf{0.3825} & 0.5293& \textbf{0.1963} & 22.75\%\\ 
\cline{2-7}
    & GVF-BCQ-0.4-CW& \textbf{2.4848} & 0.3658& \textbf{0.2953} & \textbf{0.1922} & \textbf{0.0\%} \\
    & MPC-0.4-CW& -0.7168& \textbf{0.3836} & 1.3182& 0.2095& 91.22\%\\ 
\cline{2-7}
    & GVF-BCQ-0.25-CCW & \textbf{1.5112} & \textbf{0.2473} & \textbf{0.3645} & 0.1466& \textbf{3.32\%} \\
    & MPC-0.25-CCW & 0.0225& 0.2296& 0.9565& \textbf{0.1381} & 87.92\%\\ 
\hline
\multirow{6}{*}{\begin{tabular}[c]{@{}l@{}}Complex\\ shape \end{tabular}}
    & GVF-BCQ-0.4-CCW  & \textbf{2.3501} & 0.3129& \textbf{0.2221} & \textbf{0.1817} & \textbf{0.0\%} \\
    & MPC-0.4-CCW  & 0.7172& \textbf{0.3845} & 0.6407& 0.2131& 38.94\%\\ 
\cline{2-7}
    & GVF-BCQ-0.4-CW& \textbf{2.3182} & 0.3168& \textbf{0.2317} & \textbf{0.2150} & \textbf{0.06\%} \\
    & MPC-0.4-CW& 0.4324& \textbf{0.3905} & 0.7662& 0.2264& 52.23\%\\ 
\cline{2-7}
    & GVF-BCQ-0.25-CCW & \textbf{1.9326} & \textbf{0.2472} & 0.1890& \textbf{0.1509} & 0.0\%\\
    & MPC-0.25-CCW & 1.1559& 0.2435& \textbf{0.1664} & 0.1720& 0.0\%\\
\hline
\end{tabular}
\end{table*}

\begin{table*}[t]
\centering
\caption{Comparison of jerk of GVF-BCQ method and MPC on all the test roads at different speeds and directions}
\label{table_jackal_result_gvf_vs_mpc2}
\begin{tabular}{|l|l|l|l|l|l|} 
\hline
    & Experiment & \multicolumn{1}{c|}{\begin{tabular}[c]{@{}c@{}}First \\ order\\ speed\\ jerk \textdownarrow \end{tabular}} & \multicolumn{1}{c|}{\begin{tabular}[c]{@{}c@{}}Second \\ order\\ speed \\ jerk \textdownarrow \end{tabular}} & \multicolumn{1}{c|}{\begin{tabular}[c]{@{}c@{}}First\\ order\\ steering\\ jerk \textdownarrow \end{tabular}} & \multicolumn{1}{c|}{\begin{tabular}[c]{@{}c@{}}Second\\ order\\ jerk \\ jerk \textdownarrow \end{tabular}}  \\ 
\hline
\multirow{6}{*}{\begin{tabular}[c]{@{}l@{}}Rectangle\\ shape \end{tabular}}
    & GVF-BCQ-0.4-CCW  & \textbf{0.0356}& \textbf{0.2532}  & \textbf{0.2251}  & \textbf{1.3403}  \\
    & MPC-0.4-CCW  & 0.0832  & 0.7605 & 1.2542 & 8.1963 \\ 
\cline{2-6}
    & GVF-BCQ-0.4-CW& \textbf{0.0311}& \textbf{0.2149}  & \textbf{0.1995}  & \textbf{1.1850}  \\
    & MPC-0.4-CW& 0.0944  & 0.8916 & 1.4328 & 10.9570\\ 
\cline{2-6}
    & GVF-BCQ-0.25-CCW & \textbf{0.0009}& \textbf{0.0112}  & \textbf{0.1466}  & \textbf{0.8890}  \\
    & MPC-0.25-CCW & 0.0570  & 0.5272 & 0.6384 & 3.5208 \\ 
\hline
\multirow{6}{*}{\begin{tabular}[c]{@{}l@{}}Oval\\ shape \end{tabular}}
    & GVF-BCQ-0.4-CCW  & \textbf{0.0348}& \textbf{0.2423}  & \textbf{0.2191}  & \textbf{1.4632}  \\
    & MPC-0.4-CCW  & 0.1026  & 0.9301 & 1.4119 & 8.9051 \\ 
\cline{2-6}
    & GVF-BCQ-0.4-CW& \textbf{0.0241}& \textbf{0.1638}  & \textbf{0.1674}  & \textbf{1.1451}  \\
    & MPC-0.4-CW& 0.0847  & 0.7534 & 1.3957 & 9.0432 \\ 
\cline{2-6}
    & GVF-BCQ-0.25-CCW & \textbf{0.0005}& \textbf{0.0061}  & \textbf{0.0969}  & \textbf{0.7614}  \\
    & MPC-0.25-CCW & 0.0657  & 0.6273 & 0.4830 & 3.0566 \\ 
\hline
\multirow{6}{*}{\begin{tabular}[c]{@{}l@{}}Complex\\ shape \end{tabular}}
    & GVF-BCQ-0.4-CCW  & \textbf{0.0341}& \textbf{0.2540}  & \textbf{0.2272}  & \textbf{1.5306}  \\
    & MPC-0.4-CCW  & 0.0625  & 0.5846 & 1.2133 & 8.1747 \\ 
\cline{2-6}
    & GVF-BCQ-0.4-CW& \textbf{0.0348}& \textbf{0.2339}  & \textbf{0.2240}  & \textbf{1.3911}  \\
    & MPC-0.4-CW& 0.0809  & 0.7521 & 1.2861 & 8.7905 \\ 
\cline{2-6}
    & GVF-BCQ-0.25-CCW & \textbf{0.0006}& \textbf{0.0082}  & \textbf{0.1696}  & \textbf{1.0394}  \\
    & MPC-0.25-CCW & 0.0525  & 0.4932 & 0.6457 & 3.6786 \\
\hline
\end{tabular}
\end{table*}

Details evaluations against the MPC baseline are shown in Tables \ref{table_jackal_result_gvf_vs_mpc1} and \ref{table_jackal_result_gvf_vs_mpc2}.
In our evaluation at 0.25 m/s in the counterclockwise direction, the gap between the controllers narrowed but GVF-BCQ still out-performed MPC overall.
A clear advantage of GVF-BCQ is the stability and smoothness of control achieved at the higher speeds.
Our proposed GVF-BCQ method beats the MPC in reward and was better on all tracks at both speed values without access to localization information during testing.
The reason for this can be explained by looking at the average lane centeredness of the agent.
The MPC performed as well as the GVF-BCQ method while maintaining good speed; however, it fails at keeping the vehicle in the center of the lane.
The reason may be due to a number of different factors including possible inaccuracies in the MPC forward model resulting from the friction between the wheels and the carpet in the test runs, especially at higher speeds.
The MPC suffered from many near out of lane events and had trouble staying within the lane markings of the oval road.
The GVF-BCQ method was better at controlling steering, leading to much higher average reward even though it had lower average speed at 0.4 m/s max speed.
Additionally, the GVF-BCQ method was much better in achieving smooth control.
These points are reflected in Figure \ref{fig_jackal_gvf_vs_mpc} on the rectangle test track where the MPC lane centeredness distribution is skewed to one side and its steering action distribution has two modes that are far from zero while the GVF-BCQ method distributions are more concentrated around zero.

\begin{figure*}[t]
	\centering
	\includegraphics[width=16cm]{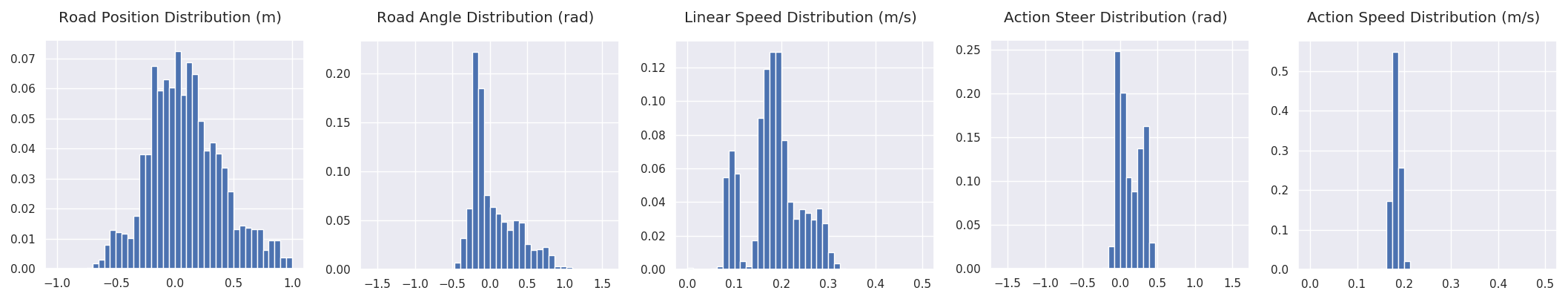}
	\caption{Distribution of lane centeredness, road angle, speed and action distribution of E2E-BCQ on the rectangle test road at 0.4 speed, counterclockwise direction}
	\label{fig_jackal_image_bcq}
\end{figure*}

\begin{figure*}[t]
	\centering
	\includegraphics[width=16cm]{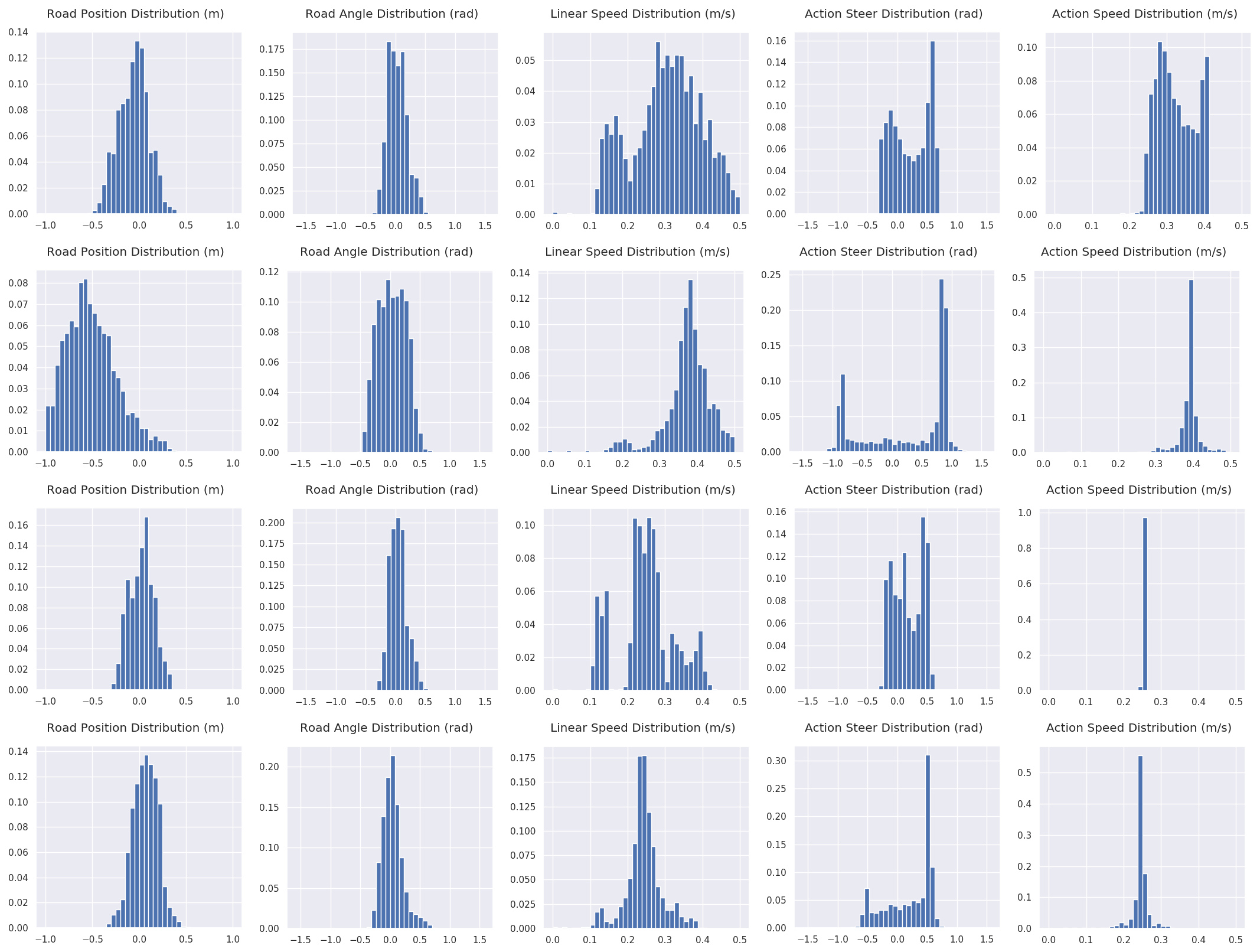}
	\caption{Distribution of lane centeredness, road angle, speed and action distribution of GVF-BCQ and MPC at 0.4 m/s and 0.25 m/s on the rectangle test track. From top to bottom: GVF-BCQ-0.4-CCW, MPC-0.4-CCW, GVF-BCQ-0.25-CCW, MPC-0.4-CCW\protect\footnotemark}
	\label{fig_jackal_gvf_vs_mpc}
\end{figure*}

In order to provide more insight into the performance of the controllers, we also investigated the distributions of $\alpha_t$, $\beta_t$, $v_t$ and $a_t$ in Figures \ref{fig_jackal_gvf_N_vs_D}, \ref{fig_jackal_image_bcq}, and \ref{fig_jackal_gvf_vs_mpc}.
\footnotetext{Note that measured vehicle speed might not be equal to speed action from the agent due to physical constraints of the environment and noises in measurement.}

\clearpage
\subsection{TORCS Experiments}
TORCS is a racing simulator used for learning to drive.
All opponent vehicles were removed for these experiments as well as roads that were sloped.
The goal of the agent is to maximize the future accumulation of the following reward: $r_t=0.0002 v_t (\cos{\beta_t} + |\alpha_t|)$ where $v_t$ is the speed of the vehicle in km/h, $\beta_t$ is the angle between the road direction and the vehicle direction, and $\alpha_t$ is the current lane centeredness.
Termination occurs when either the agent leaves the lane or the maximum number of steps has been reached (1200 steps = 120 seconds) triggering a reset of the environment.
Upon reset, a priority sampling method is used during training to select the next road to train on.
The probability of sampling road $i$ during a reset is given by 

\begin{equation}
\frac{e^{-\frac{n_i}{\kappa}}}{\sum_{j=1}^{N}{e^{-\frac{n_j}{\kappa}}}}
\label{eq_track_sample}
\end{equation}

where $n_i$ is the number of steps that the agent was able to achieve the last time the road was sampled and $\kappa$ controls the spread of the distribution.
A value of $\kappa=\frac{1}{N}\sum_{j=1}^{N}{n_j}$ was found to perform well.
The initial probabilities are equal for all roads.
This improved the efficiency in learning for all learned methods.

The TORCS environment was modified to provide higher resolution images in grayscale rather than RGB with most of the image above the horizon cropped out of the image.
The grayscale images were 128 pixels wide by 64 pixels high.
This allowed the agent to see more detail farther away which is very helpful in making long term predictions and is beneficial to both policy gradient methods and predictive learning.

\subsubsection{Training}
Two main learning algorithms are compared, along with their variants:  our proposed GVF-DDPG (general value functions with deep deterministic policy gradient) and end-to-end DDPG.
The parameters used to train the methods will be described in more detail here.

\textbf{GVF-DDPG Training:}
Exploration followed the same approach as \cite{lillicrap2016} where an Ornstein Uhlenbeck process \cite{uhlenbeck1930} was used to explore the road; the parameters of the process ($\theta=1.0$, $\sigma=0.1$, $dt=0.01$) were tuned to provide a gradual wandering behavior on the road without excessive oscillations in the action.
This improved the learning of the off-policy predictions for GVF-DDPG since the behavior policy $\mu(a|s)$ was closer to the target policy $\tau(a|s)$ of the predictions.

\begin{figure}[t]
	\centering
	\begin{subfigure}[b]{4cm}
		\includegraphics[width=4cm]{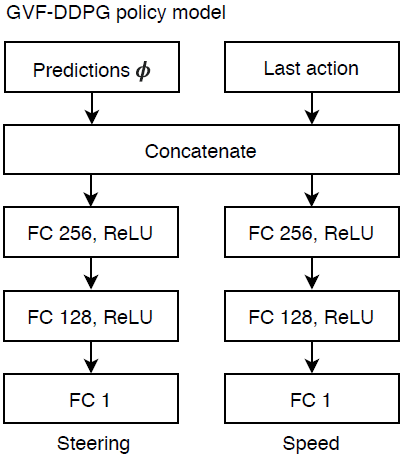}
		\caption{Actor Model}
		\label{fig_gvfddpg_actor_model}
	\end{subfigure}
	\begin{subfigure}[b]{4cm}
		\includegraphics[width=4cm]{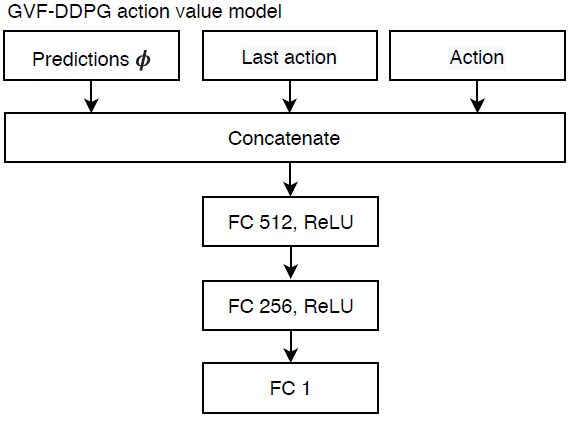}
    	\caption{Critic Model}
    	\label{fig_gvfddpg_critic_model}
	\end{subfigure}
	\caption{Neural network models for GVF-DDPG (a) Actor, and (b) Critic}
	\label{fig_gvfddpg_actor_critic_models}
\end{figure}

The GVF-DDPG approach learned 8 predictions:  4 predictions of lane centeredness $\alpha$, and 4 predictions of road angle $\beta$.
Each of the 5 predictions had different values of $\gamma$ for different temporal horizons:  0.0, 0.5, 0.9, 0.95, 0.97.
This allowed the agent to predict how the road will turn in the future providing the necessary look ahead information for the agent to control effectively.
The GVF predictors shared the same deep convolutional neural network as used on the Jackal robot where the convolutional layers were identical to the architecture in Figure \ref{fig_feature_extractor_model} followed by three fully connected layers of 512, 384 and 8 outputs, respectively as shown in Figure \ref{fig_prediction_model}.
The behavior estimator $\mu(a|s)$ was identical with the one used on the Jackal robot in Figure \ref{fig_behavior_model}.
The models for actor and critic models for GVF-DDPG are given in Figures \ref{fig_gvfddpg_actor_model} and \ref{fig_gvfddpg_critic_model}.
Actions produced by the actor network was clipped to the range $[-1.0, 1.0]$.
A linear transformation was applied to the target speed action to change the range of values from $[-1.0, 1.0]$ to $[0.5, 1.0]$.
The steering and target speed input into both policy and critic networks were normalized to range $[-1.0, 1.0]$.

A replay buffer of size 100,000 was used with a warmup of 10,000 samples.
In order to not bias the replay buffer, the last layers of the actor network were initialized with a uniform distribution $[-1e^{-3}, 1e^{-3}]$ for the weight and 0 for the bias.
The learning rates for the actor network, critic network, predictor network and behavior policy network were $1e^{-6}$, $1e^{-4}$, $1e^{-4}$, and $1e^{-4}$ respectively.
Target networks \cite{lillicrap2016} were used for the critic and actor networks with $\tau=0.001$ in order to make the bootstrapped prediction of the action-values more stable.
However, target networks were not necessary for the GVF predictions or the behavior policy estimation.
The reward for was scaled by 0.0002 to scale the action-values to fall within the range $[-1.0,1.0]$.

\textbf{Baseline DDPG Training:}
The two DDPG (deep deterministic policy gradient) \cite{lillicrap2016} baselines were trained nearly identically where the only difference was the information provided in the observation.
The first method called DDPG-Image is a vision-based approach where the image, current speed, and last action are provided to the agent; the only information available to the agent about the road is supplied via images.
The second agent called DDPG-LowDim includes lane centeredness $\alpha$ and road angle $\beta$ as part of the agent's state.
The purpose was to understand the value of this information in learning when supplied as a cumulant in GVF-DDPG during training only or supplied an input to the actor and critic during training \textit{and} testing.
It should be noted that DDG-LowDim was the only learned approach that used $\alpha$ and $\beta$ was inputs to the actor and critic networks during testing, whereas GVF-DDPG and DDPG-Image did not have access to this information during testing.

\begin{figure}[t]
	\centering
	\includegraphics[width=6cm]{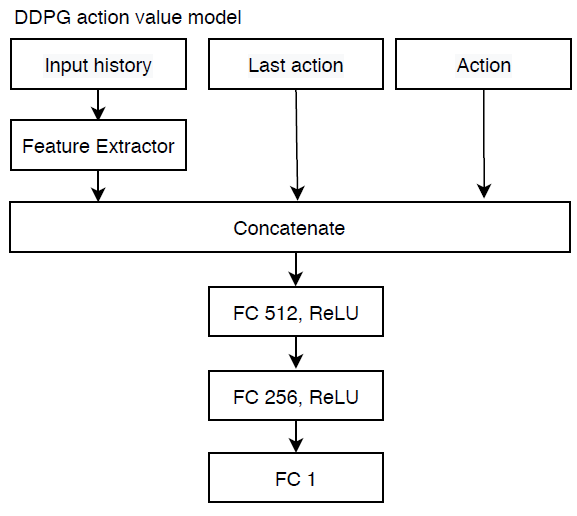}
	\caption{Model of the DDPG critic network $Q(s,a)$.}
	\label{fig_ddpg_critic_model}
\end{figure}

\begin{figure}[t]
	\centering
	\includegraphics[width=4cm]{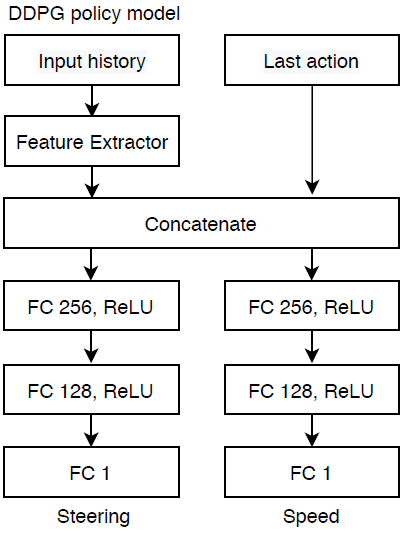}
	\caption{Model of the DDPG actor network $\pi(s)$.}
	\label{fig_ddpg_actor_model}
\end{figure}

The network architectures for the DDPG actor and critic are given in Figures \ref{fig_ddpg_actor_model} and \ref{fig_ddpg_critic_model} respectively; they share the architecture for the feature extractor given in Figure \ref{fig_feature_extractor_model}.


The training setup for DDPG was the same as GVF-DDPG unless noted otherwise.
An Ornstein Uhlenbeck process \cite{uhlenbeck1930} was used to explore the road ($\theta=1.0$, $\sigma=0.1$, and $dt=0.01$).
Experimentally, it was found that the exploration parameters did not affect the learning performance of DDPG that much.
Target networks were used with $\tau=0.001$.
The learning rates of the critic and actor networks were the same as those of GVF-DDPG.

\subsubsection{Experimental Results}

\begin{figure}[t]
	\centering
	\begin{subfigure}[b]{8cm}
		\includegraphics[width=8cm]{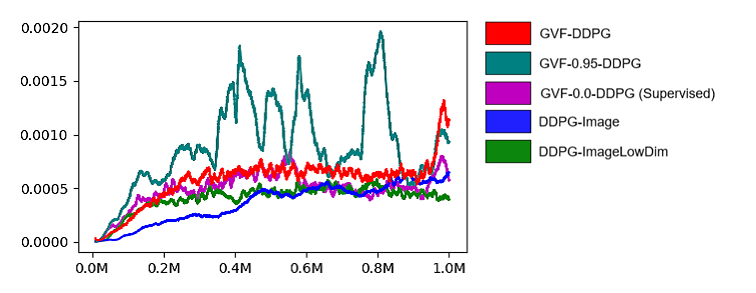}
		\caption{Critic}
	\end{subfigure}
	\\
	\begin{subfigure}[b]{4cm}
		\includegraphics[width=4cm]{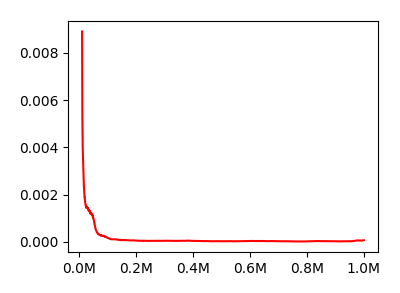}
		\caption{Predictors}
	\end{subfigure}
	\begin{subfigure}[b]{4cm}
		\includegraphics[width=4cm]{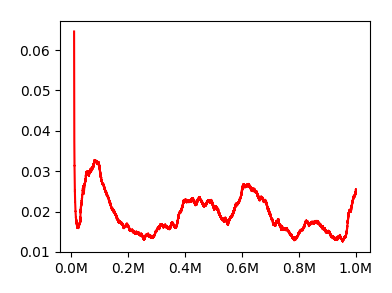}
		\caption{Behavior}
	\end{subfigure}
	\caption{Learning curves for (a) Q-values of the DDPG agents, (b) mean squared TD (temporal difference) errors of the GVF predictors, and (c) MSE of the behavior model estimator}
	\label{fig_learning_curve}
\end{figure}

\begin{figure}[t]
	\centering
	\includegraphics[width=8cm]{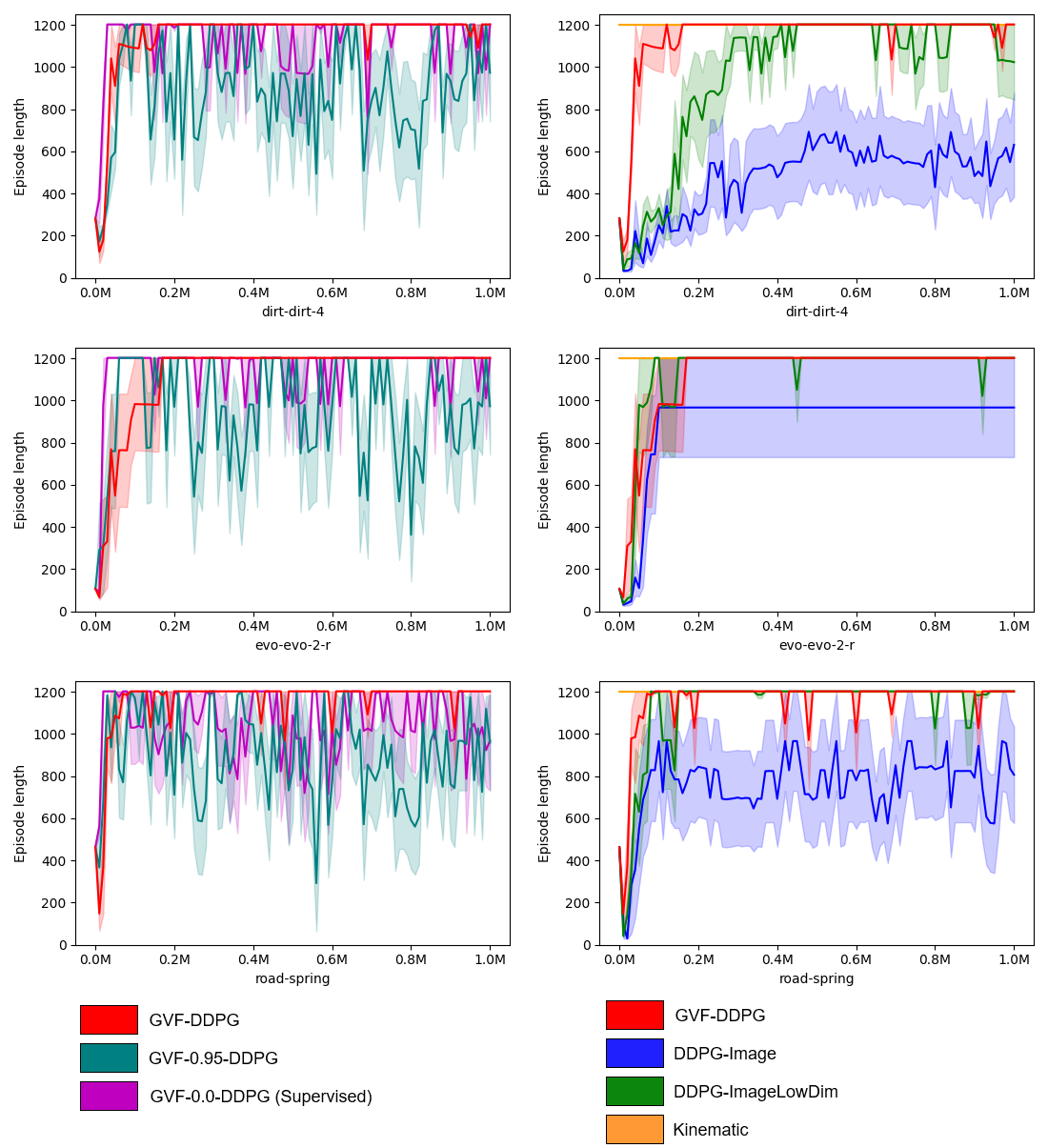}
	\caption{Mean episode length during training for dirt-dirt-4, evo-evo-2 and road-spring}
	\label{fig_episode_length}
\end{figure}

\begin{figure}[t]
	\centering
	\includegraphics[width=8cm]{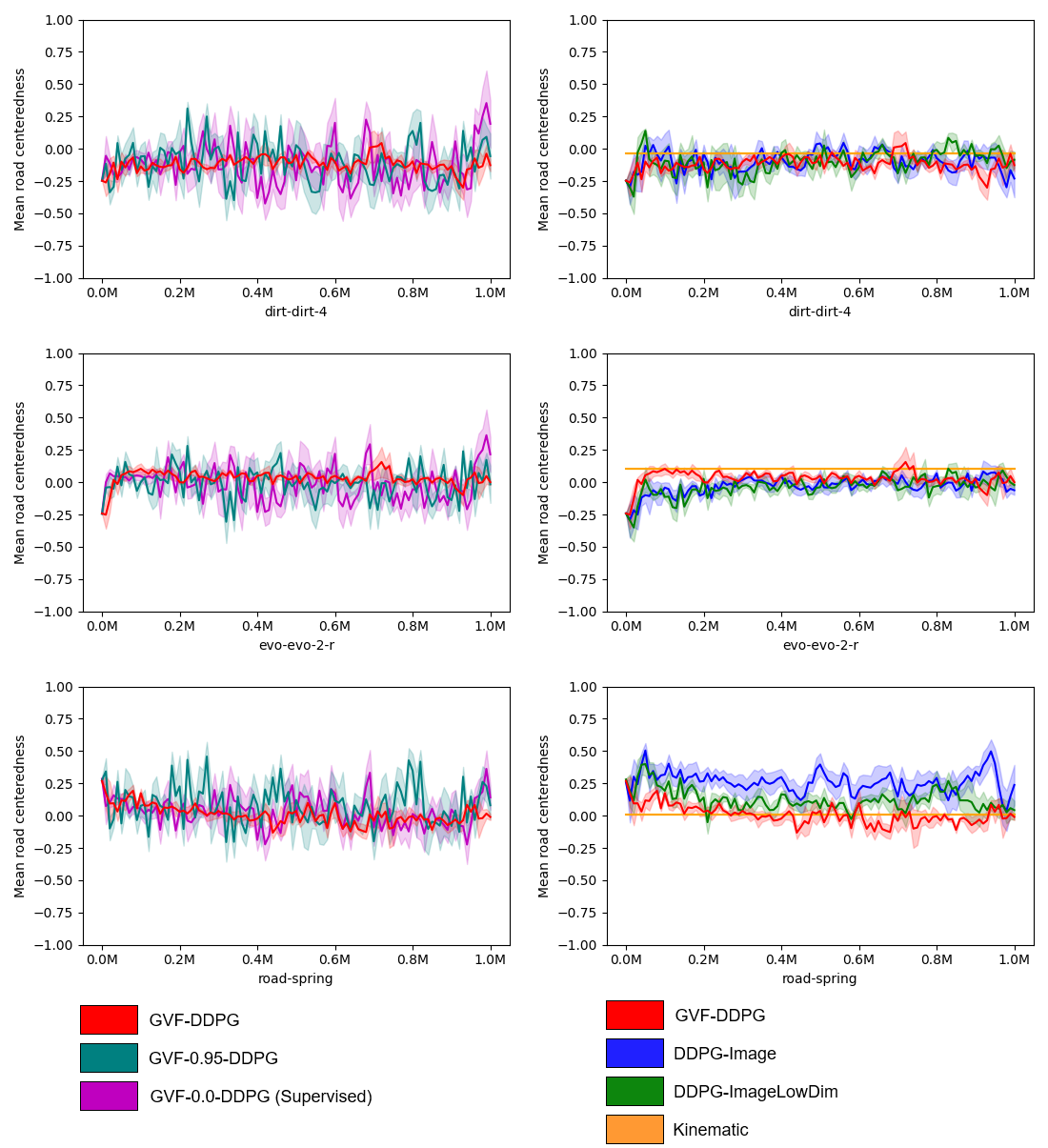}
	\caption{Mean lane centeredness during training for dirt-dirt-4, evo-evo-2 and road-spring}
	\label{fig_road_centeredness}
\end{figure}

\begin{figure}[t]
	\centering
	\includegraphics[width=8cm]{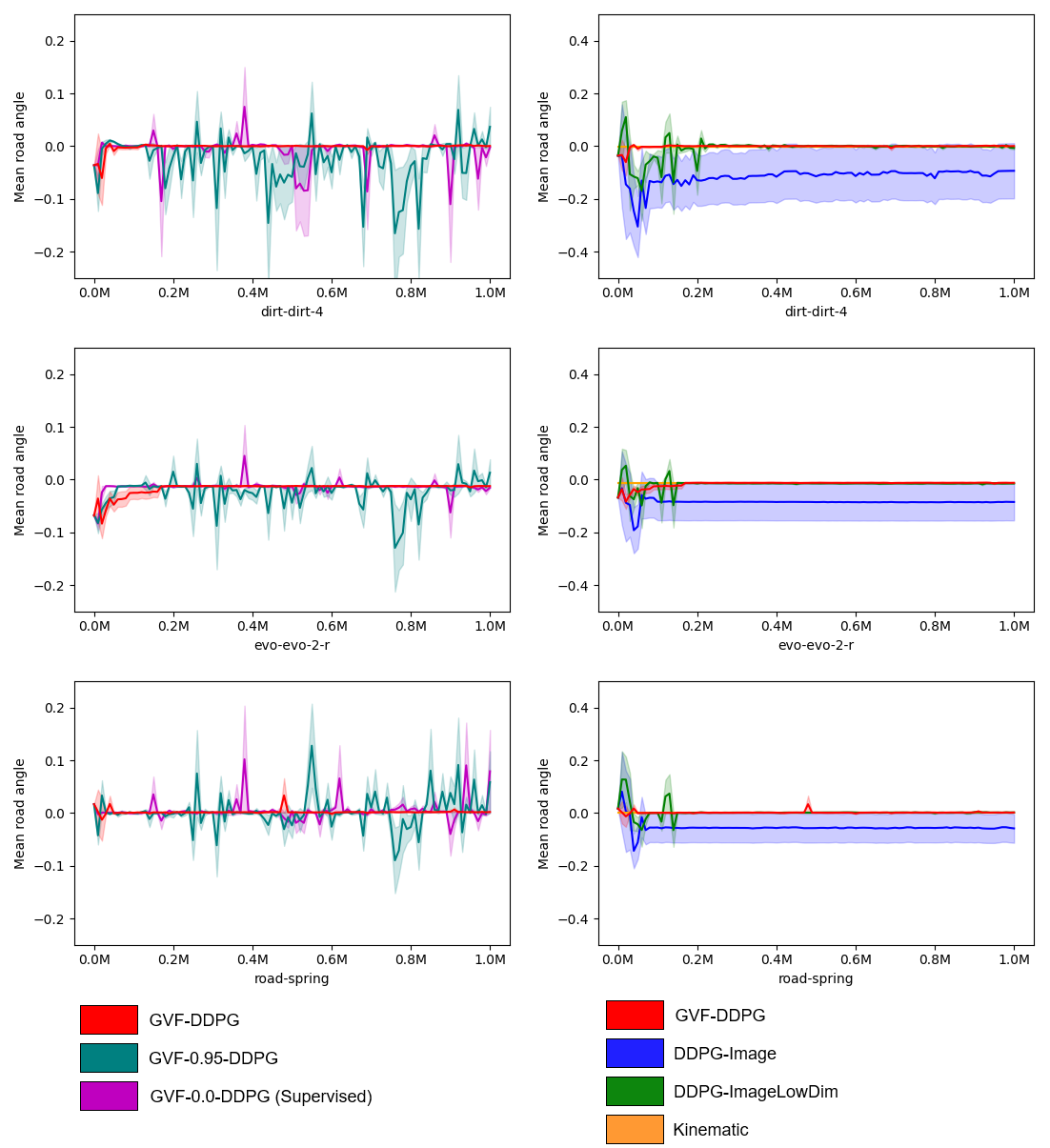}
	\caption{Mean road angle during training for dirt-dirt-4, evo-evo-2 and road-spring}
	\label{fig_road_angle}
\end{figure}

\begin{figure}[t]
	\centering
	\includegraphics[width=8cm]{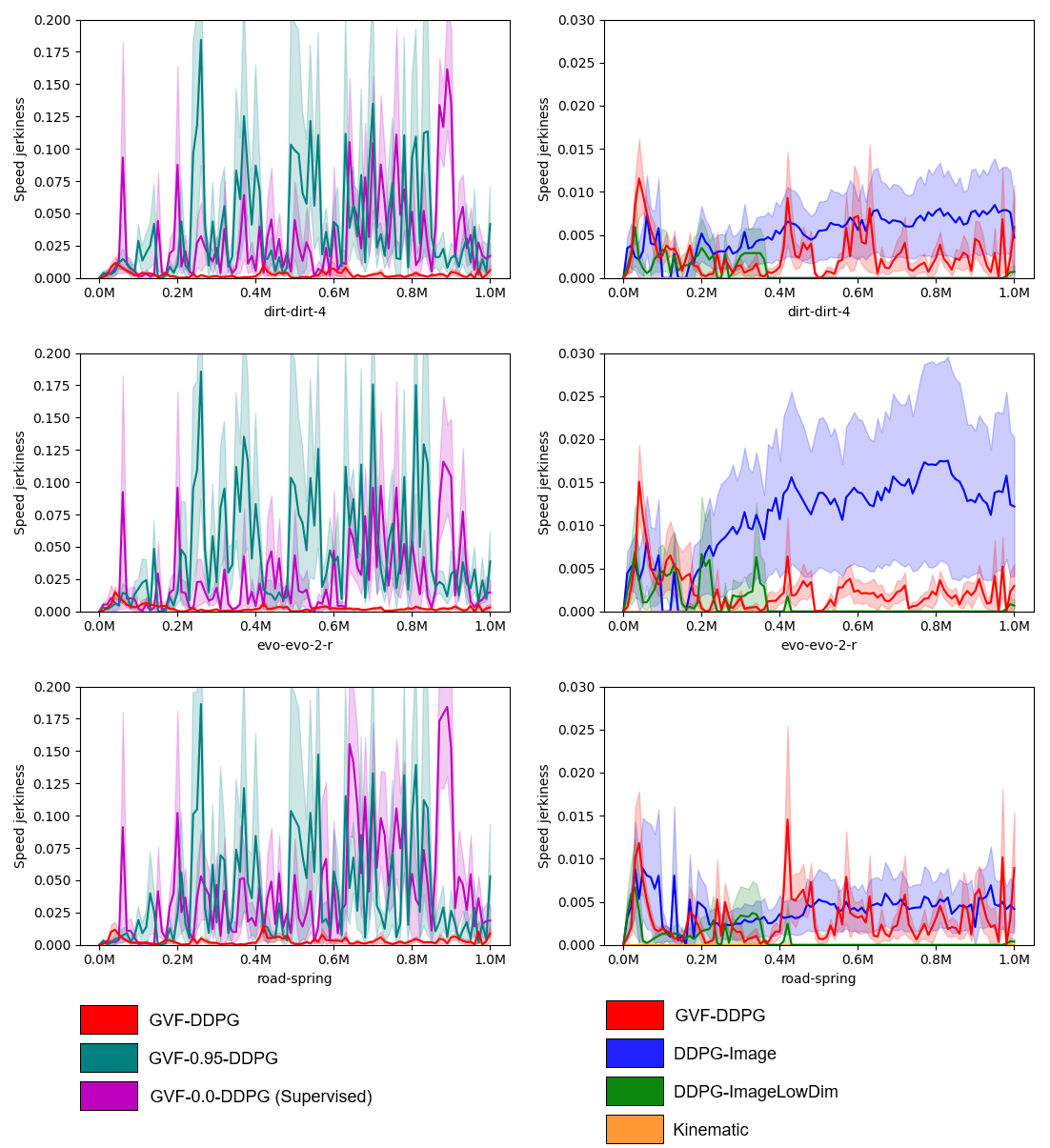}
	\caption{Standard deviation of the change in target speed action during training for dirt-dirt-4, evo-evo-2 and road-spring}
	\label{fig_delta_action_speed_std}
\end{figure}

\begin{figure}[t]
	\centering
	\includegraphics[width=8cm]{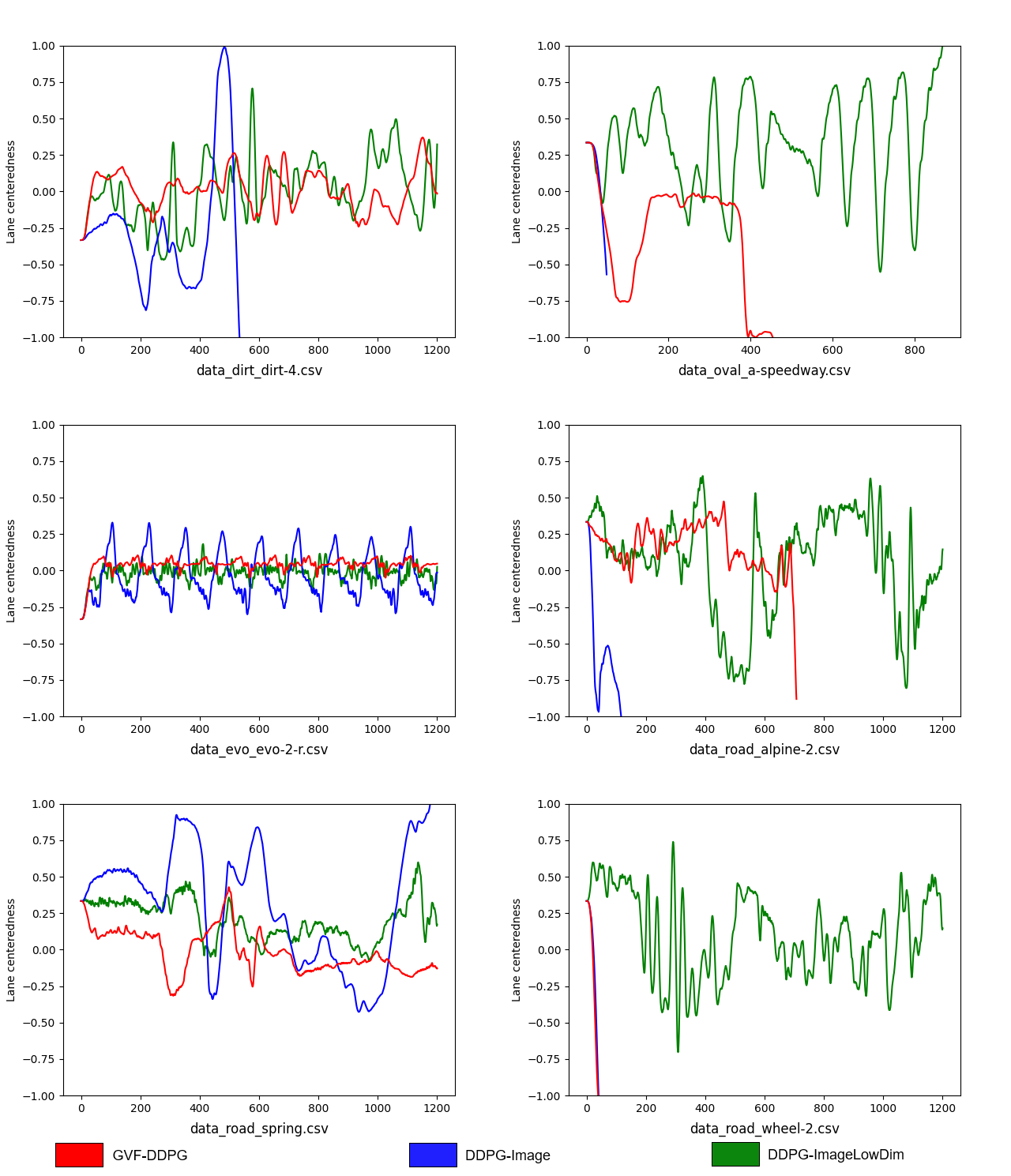}
	\caption{The lane centeredness position on the (a) alpine-2, (b) evo-2-r, (c) dirt-4, (d) wheel-2, (e) spring, and (f) a-speedway roads in TORCS.}
\label{fig_track_lane_centeredness}
\end{figure}

The experimental results were averaged over 5 runs and mean and standard deviations plotted based on performance measured on the test roads during training.
The learning curves for the critic networks for each of the DDPG agents, as well as learning curves for GVF predictions and behavior estimation for GVF-DDPG are shown in Figure \ref{fig_learning_curve}.
The average episode length is shown in Figure \ref{fig_episode_length}.
The average lane centeredness and road angle during each episode are plotted in Figures \ref{fig_road_centeredness} and \ref{fig_road_angle}.
We can see how the GVF-DDPG with predictions over multiple time scales is able to maintain lane centeredness and road angle better than GVF-DDPG with myopic prediction ($\gamma=0.0$) and future predictions with only $\gamma=0.95$.
The average lane centeredness and road angle is not substantially different though amoung the learned methods; however, DDPG-Image struggles a bit largely due to instability in learning as the high variance is due to some failed runs where no learning occurs.
Figure \ref{fig_delta_action_speed_std} shows the standard deviation in the change in the target speed action at each time step across an episode on each test road; this measures the jerkiness of the speed controller.
GVF-DDPG and DDPG-LowDim are both able to control speed comfortably since the jerkiness is low.
Finally, Figure \ref{fig_track_lane_centeredness} shows the lane centeredness on all six of the test roads during a final evaluation after training was completed.
All six roads in the test set were challenging but the test roads a-speedway, alpine-2, and wheel-2 were especially challenging because the image of the roads were too different from the training roads.
Nevertheless, on the dirt-4, evo-2-r, and spring roads, the lane centeredness of the the methods was quite good for all learned methods except for DDPG-Image.
Note that while DDPG-LowDim performs well in learning to steer and control speed with lane centeredness and road angle, it required that information to learn to steer the vehicle which may be expensive or prohibitive to obtain in all situations such as in GPS-denied locations or locations where there is no map or it is out of date.

\clearpage
\subsection{Predictive Learning Algorithms}

The algorithm used for learning counterfactual predictions (GVFs) online through interaction with an environment is given in Algorithm \ref{alg_gvf_offpolicy_train_without_mu_online}.
In important distinction of this algorithm is that the distribution of the behavior policy used to collect the data does not need to be known.

\begin{algorithm}
\caption{Online Counterfactual GVF training algorithm with unknown $\mu(a|s)$}
\label{alg_gvf_offpolicy_train_without_mu_online}
\begin{algorithmic}[1]
\State Initialize $\phi^\tau(s)$, $g(a,s)$, $\eta(a|s)$, and replay memory $D$
\State Observe initial state $s_0$
\For{$t=0$,$T$}
  \State Sample action $a_t$ from unknown $\mu(a_t|s_t)$
  \State Execute action $a_t$ and observe state $s_{t+1}$
  \State Compute cumulant $c_{t+1}=c(s_t,a_t,s_{t+1})$
  \State Compute continuation $\gamma_{t+1}=\gamma(s_t,a_t,s_{t+1})$
  \State Estimate behavior density value $\hat{\mu}(a_t|s_t)=\frac{g(a_t,s_t)}{1-g(a_t,s_t)}\eta(a_t|s_t)$
  \State Estimate importance sampling ratio $\rho_t=\frac{\tau(a_t|s_t)}{\hat{\mu}(a_t|s_t)}$
  \State Store transition $(s_t,a_t,c_{t+1},\gamma_{t+1},s_{t+1},\rho_t)$ in $D$
  \State Compute average importance sampling ratio in replay buffer $D$ of size $n$ with $\bar{\rho}=\frac{1}{n}\sum_{j=1}^{n}{\rho_j}$
  \State Sample random minibatch $A$ of transitions $(s_i,a_i,c_{i+1},\gamma_{i+1},s_{i+1})$ from $D$ according to probability $\frac{\rho_i}{\sum_{j=1}^{n}\rho_j}$
  \State Compute $y_i = c_{i+1} + \gamma_{i+1} \phi^\tau(s_{i+1};\hat{\theta})$ for minibatch $A$ for most recent parameters $\hat{\theta}$
  \State Update parameters $\theta$ using gradient descent on \eqref{eq_td_loss} with gradient \eqref{eq_td_ir_gradient} over the minibatch $A$
  \State Sample random minibatch $B$ of state action pairs $(s_i,a_i)$ from $D$ according to a uniform probability and assign label $z=1$ to each pair
  \State Randomly select half the samples in the minibatch $B$ replacing the action with $a_t \sim \eta(a|s)$ and label with $z=0$ and storing the updated samples in $\hat{B}$
  \State Update behavior discriminator $g(a,s)$ with labels $z$ in the modified minibatch $\bar{B}$ using binary cross-entropy loss
\EndFor
\end{algorithmic}
\end{algorithm}

With minor modifications, an offline version of the algorithm can be derived.
This algorithm learns by reading the data in sequence and populating a replay buffer just as it would in online learning; the only difference is that the offline algorithm returns the action taken in the data.
This allows the same algorithm and code for learning counterfactual predictions (GVF) to be used in either online or offline learning settings.

\begin{algorithm}
\caption{Offline Counterfactual GVF training algorithm with unknown $\mu(a|s)$}
\label{alg_gvf_offpolicy_train_without_mu_offline}
\begin{algorithmic}[1]
\State Initialize $\phi^\tau(s)$, $g(a,s)$, $\eta(a|s)$, and replay memory $D$,
\State Obtain the first state in the data file $s_0$
\For{$t=0$,$T$}
  \State Obtain action $a_t$ recorded in the data file that sampled from an unknown $\mu(a_t|s_t)$
  \State Obtain next state $s_{t+1}$ from the data file
  \State Compute cumulant $c_{t+1}=c(s_t,a_t,s_{t+1})$
  \State Compute continuation $\gamma_{t+1}=\gamma(s_t,a_t,s_{t+1})$
  \State Estimate behavior density value $\hat{\mu}(a_t|s_t)=\frac{g(a_t,s_t)}{1-g(a_t,s_t)}\eta(a_t|s_t)$
  \State Estimate importance sampling ratio $\rho_t=\frac{\tau(a_t|s_t)}{\hat{\mu}(a_t|s_t)}$
  \State Store transition $(s_t,a_t,c_{t+1},\gamma_{t+1},s_{t+1},\rho_t)$ in $D$
  \State Compute average importance sampling ratio in replay buffer $D$ of size $n$ with $\bar{\rho}=\frac{1}{n}\sum_{j=1}^{n}{\rho_j}$
  \State Sample random minibatch $A$ of transitions $(s_i,a_i,c_{i+1},\gamma_{i+1},s_{i+1})$ from $D$ according to probability $\frac{\rho_i}{\sum_{j=1}^{n}\rho_j}$
  \State Compute $y_i = c_{i+1} + \gamma_{i+1} \phi^\tau(s_{i+1};\hat{\theta})$ for minibatch $A$ for most recent parameters $\hat{\theta}$
  \State Update parameters $\theta$ using gradient descent on \eqref{eq_td_loss} with gradient \eqref{eq_td_ir_gradient} over the minibatch $A$
  \State Sample random minibatch $B$ of state action pairs $(s_i,a_i)$ from $D$ according to a uniform probability and assign label $z=1$ to each pair
  \State Randomly select half the samples in the minibatch $B$ replacing the action with $a_t \sim \eta(a|s)$ and label with $z=0$ and storing the updated samples in $\hat{B}$
  \State Update behavior discriminator $g(a,s)$ with labels $z$ in the modified minibatch $\bar{B}$ using binary cross-entropy loss
\EndFor
\end{algorithmic}
\end{algorithm}

\end{document}